\documentclass{article}

    \PassOptionsToPackage{numbers, compress}{natbib}

\usepackage[final]{neurips_2024}

\usepackage[utf8]{inputenc} 
\usepackage[T1]{fontenc}    
\usepackage{tikz}
\usepackage{multirow}
\usepackage[colorlinks,
    colorlinks=true, 
    linkcolor=blue, 
    filecolor=blue, 
    citecolor=blue,       
    urlcolor=blue]{hyperref}       
\usepackage{url} 
\usepackage{enumitem} 
\usepackage{booktabs}       
\usepackage{amsfonts}       
\usepackage{amsmath}
\usepackage{amsthm}
\usepackage{amssymb}
\usepackage{nicefrac}       
\usepackage{microtype}      
\usepackage{xcolor}         
\usepackage[font=small,labelfont=bf]{caption}
\usepackage{enumitem}
\usepackage{wrapfig}
\usepackage{listings}
\usepackage{caption}
\usepackage{amsmath}
\usepackage{mathtools}
\usepackage{subcaption}
\usepackage[hybrid]{markdown}
\usepackage{graphicx}
\usepackage{colortbl}

\usepackage[normalem]{ulem}
\usepackage{xspace}
\usepackage{float}
\usepackage{tabularx}
\usepackage{booktabs}
\usepackage[normalem]{ulem}
\useunder{\uline}{\ul}{}

\usepackage[toc,page,header]{appendix}
\usepackage{minitoc}

\newcommand\numberthis{\addtocounter{equation}{1}\tag{\theequation}}

\def\mistral{Pixtral~12B\xspace}

\newcommand{\rope}{\textsc{RoPE-2D}}
\newcommand\inner[2]{\langle #1, #2 \rangle}

\title{\mistral}

\begin{document}

\part{} 

\maketitle

\begin{center}
\vspace{-20pt}
\centering
\includegraphics[width=0.8\linewidth,keepaspectratio]{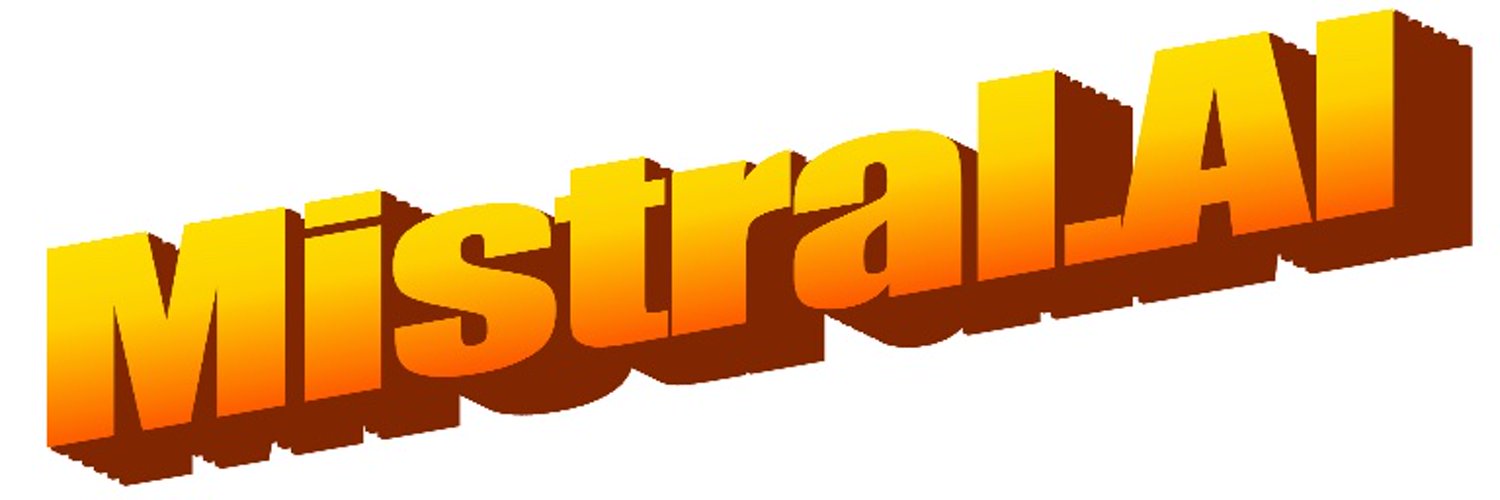}
\end{center}

\begin{abstract}
We introduce \mistral, a 12--billion-parameter multimodal language model.
\mistral is trained to understand both natural images and documents, achieving leading performance on various multimodal benchmarks, surpassing a number of larger models. 
Unlike many open-source models, Pixtral is also a cutting-edge text model for its size, and does not compromise on natural language performance to excel in multimodal tasks. 
Pixtral uses a new vision encoder trained from scratch, which allows it to ingest images at their natural resolution and aspect ratio. This gives users flexibility on the number of tokens used to process an image. Pixtral is also able to process any number of images in its long context window of 128K tokens.
Pixtral 12B substanially outperforms other open models of similar sizes (Llama-3.2 11B \& Qwen-2-VL 7B). It also outperforms much larger open models like Llama-3.2 90B while being 7x smaller.
We further contribute an open-source benchmark, MM-MT-Bench, for evaluating vision-language models in practical scenarios, and provide detailed analysis and code for standardized evaluation protocols for multimodal LLMs. \mistral is released under Apache 2.0 license.
\\

\textbf{Webpage:} \url{https://mistral.ai/news/pixtral-12b/} \\
\textbf{Inference code:} \url{https://github.com/mistralai/mistral-inference/} \\
\textbf{Evaluation code:} \url{https://github.com/mistralai/mistral-evals/}

\end{abstract}




\section{Introduction}

\begin{figure}[h]
\centering
\includegraphics[width=0.49\textwidth]{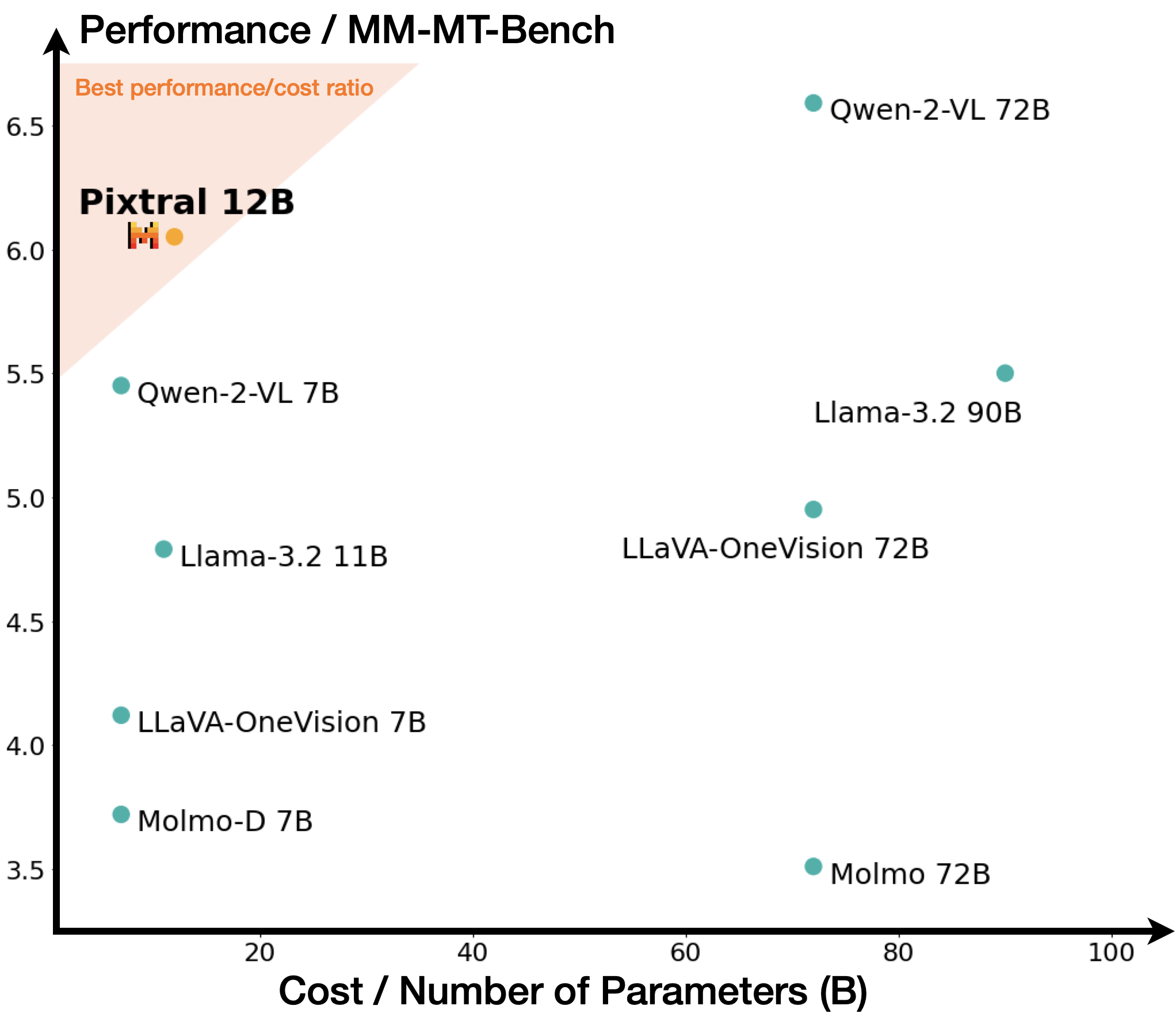}
\includegraphics[width=0.49\textwidth]{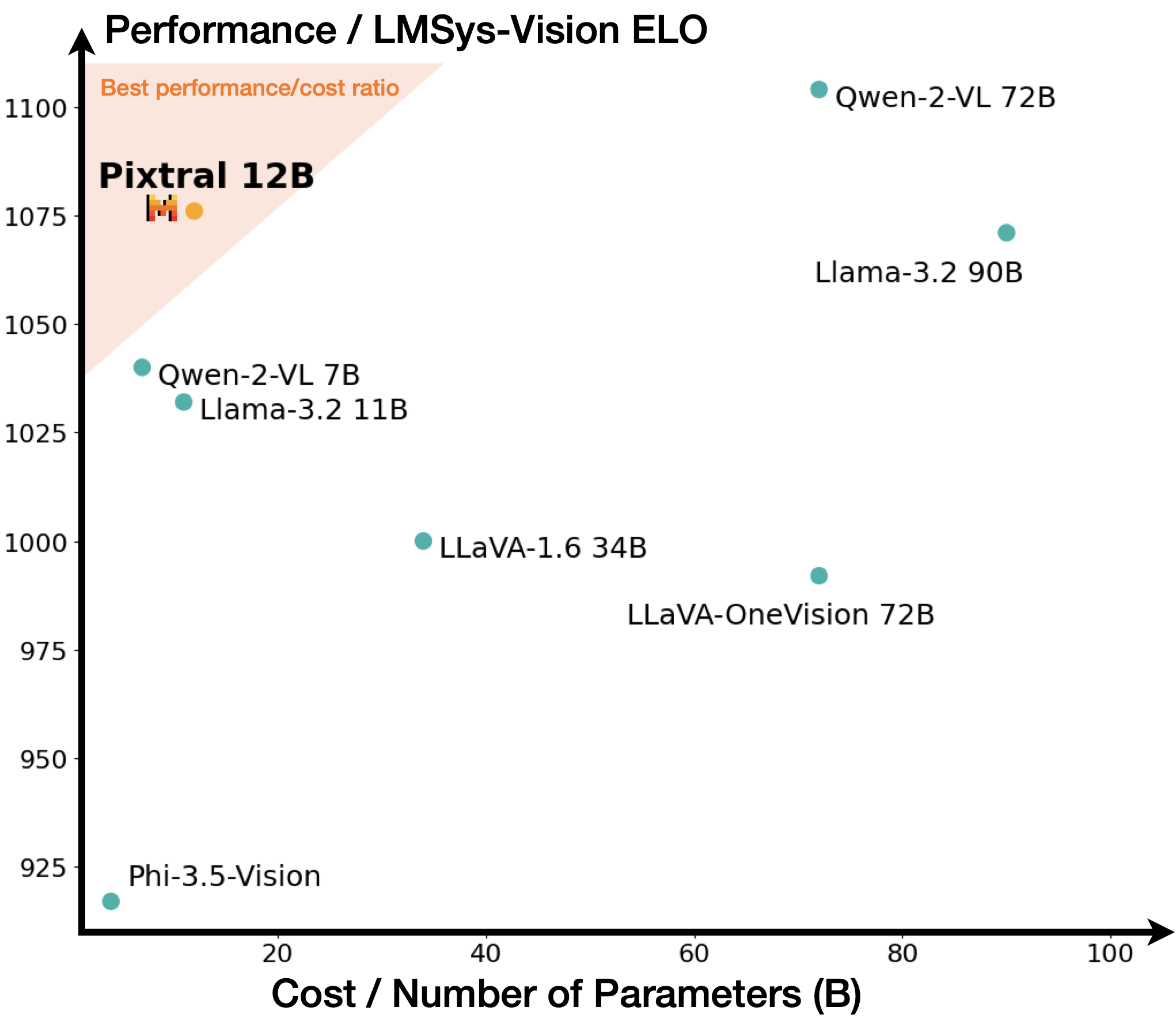}
\caption{
\small \textbf{Pixtral Performance.} 
Pixtral outperforms all open-models within its weight class on multimodal tasks by a substantial margin.
\textit{Left:} Performance on MM-MT-Bench, a new multimodal, multiturn, instruction following benchmark designed to reflect real world usage of multimodal language models.
\textit{Right:} Performance on the public LMSys leaderboard (Vision arena, October 2024). 
}
\label{fig:lmsys}
\end{figure}

This paper describes Pixtral 12B, a multimodal language model trained to understand both images and text, released with open weights under an Apache 2.0 license.
Pixtral is an instruction tuned model which is pretrained on large scale interleaved image and text documents, and hence is capable of multi-turn, multi-image conversation.

\looseness=-1 Pixtral comes with a new vision encoder which is trained with a novel \textsc{RoPE-2D} implementation, allowing it to process images at their native resolution and aspect ratio. 
In this way, the model can flexibly process images at low resolution in latency-constrained settings, while processing images at high resolution when fine-grained reasoning is required.

When compared against models of a similar size in the same evaluation setting, we find that Pixtral delivers strong multimodal reasoning capabilities without sacrificing text-only reasoning performance.
For instance, our model matches or exceeds the performance of models like Qwen2-VL 7B~\cite{wang2024qwen2vlenhancingvisionlanguagemodels} and Llama-3.2 11B~\cite{dubey2024llama3} on popular multimodal benchmarks like MMMU~\cite{yue2023mmmu} and MathVista~\cite{lu2023mathvista}, while outperforming most open-source models on popular text-only tasks like MATH~\cite{hendrycks2021measuring} and HumanEval~\cite{zhong2023agieval}. Pixtral even outperforms much larger models like Llama-3.2 90B~\cite{dubey2024llama3}, as well as closed models such as Claude-3 Haiku~\cite{claude3} and Gemini-1.5 Flash 8B~\cite{reid2024gemini}, on multimodal benchmarks.

During evaluation of Pixtral and the baselines, we found that evaluation protocols for multimodal language models is not standardized, and that small changes in the setup can dramatically change the performance of some models. 
We provide thorough analysis of our experience in re-evaluating vision-language models under a common evaluation protocol. 

Specifically, we identify two issues with evaluation:

\textbullet~\textbf{Prompts:} Several benchmarks have default prompts which are under-specified, and dramatically reduce the performance of leading closed source models~\cite{openai2023gpt,claude3} compared to reported figures.

\textbullet~\textbf{Evaluation Metrics:} The official metrics typically require \textit{exact match}, which score model generations as correct only if they exactly match the reference answer. However, this metric penalizes answers which are substantively correct but in a slightly different format (\textit{e.g.}, \texttt{"6.0"} vs \texttt{"6"}).

To alleviate these issues, we propose `Explicit' prompts that explicitly specify the format required by the reference answer. 
We further analyze the impact of flexible parsing for various models, releasing the evaluation code and prompts in an effort to establish fair and standardized evaluation protocols\footnote{\url{https://github.com/mistralai/mistral-evals/}}.

Moreover, while current multimodal benchmarks mostly evaluate short-form or multiple-choice question answering given an input image, they do not fully capture a model's utility for practical use cases (\textit{e.g.} in a multi-turn, long-form assistant setting). 
To address this, we open-source a novel multimodal, multi-turn evaluation: MM-MT-Bench\footnote{\url{https://huggingface.co/datasets/mistralai/MM-MT-Bench}}. 
We find that performance on MM-MT-Bench correlates highly with ELO rankings on the LMSys Vision Leaderboard.

Pixtral excels at multimodal instruction following, surpassing comparable open-source models on the MM-MT-Bench benchmark (see Figure~\ref{fig:lmsys}). Based on human preferences on the LMSys Vision Leaderboard, Pixtral 12B is currently the highest ranked Apache 2.0 model, substantially outperforming other open-models such Llama-3.2 11B~\cite{dubey2024llama3} and Qwen2-VL 7B~\cite{wang2024qwen2vlenhancingvisionlanguagemodels}. It even ranks higher than several closed models such as Claude-3 Opus \& Claude-3 Sonnet~\cite{claude3}, and several larger models such as Llama-3.2 90B~\cite{dubey2024llama3}.

\newpage

\section{Architectural details}

\begin{wrapfigure}{r}{0.4\textwidth}
\center
\small
\vspace{-25pt}
\begin{tabular}{lrr}
\toprule
    \textbf{Parameters} & \textbf{Decoder} & \textbf{Encoder} \\
\midrule
    \texttt{dim} & 5120 & 1024 \\
    \texttt{n\_layers} & 40 & 24 \\
    \texttt{head\_dim}  & 128 & 64 \\
    \texttt{hidden\_dim} & 14336 & 4096 \\
     \texttt{n\_heads} & 32 & 16 \\
    \texttt{n\_kv\_heads}  & 8 & 16 \\
    \texttt{context\_len} & 131072 & 4096 \\
    \texttt{vocab\_size} & 131072 & - \\
    \texttt{patch\_size} & - & 16 \\
    \bottomrule
\end{tabular}
\captionof{table}{\small Decoder and encoder parameters.}
\label{tab:param}
\vspace{-30pt}
\end{wrapfigure}

\mistral is based on the transformer architecture~\cite{vaswani2017attention}, and consists of a \textit{multimodal decoder} to perform high-level reasoning, and a \textit{vision encoder} to allow the model to ingest images.
The main parameters of the model are summarized in Table~\ref{tab:param}.

\subsection{Multimodal Decoder}

\mistral is built on top of Mistral Nemo 12B~\cite{mistralnemo12b}, a 12-billion parameter decoder-only language model that achieves strong performance across a range of knowledge and reasoning tasks.

\subsection{Vision Encoder}

\begin{figure} 
\centering
\includegraphics[width=0.97\linewidth]{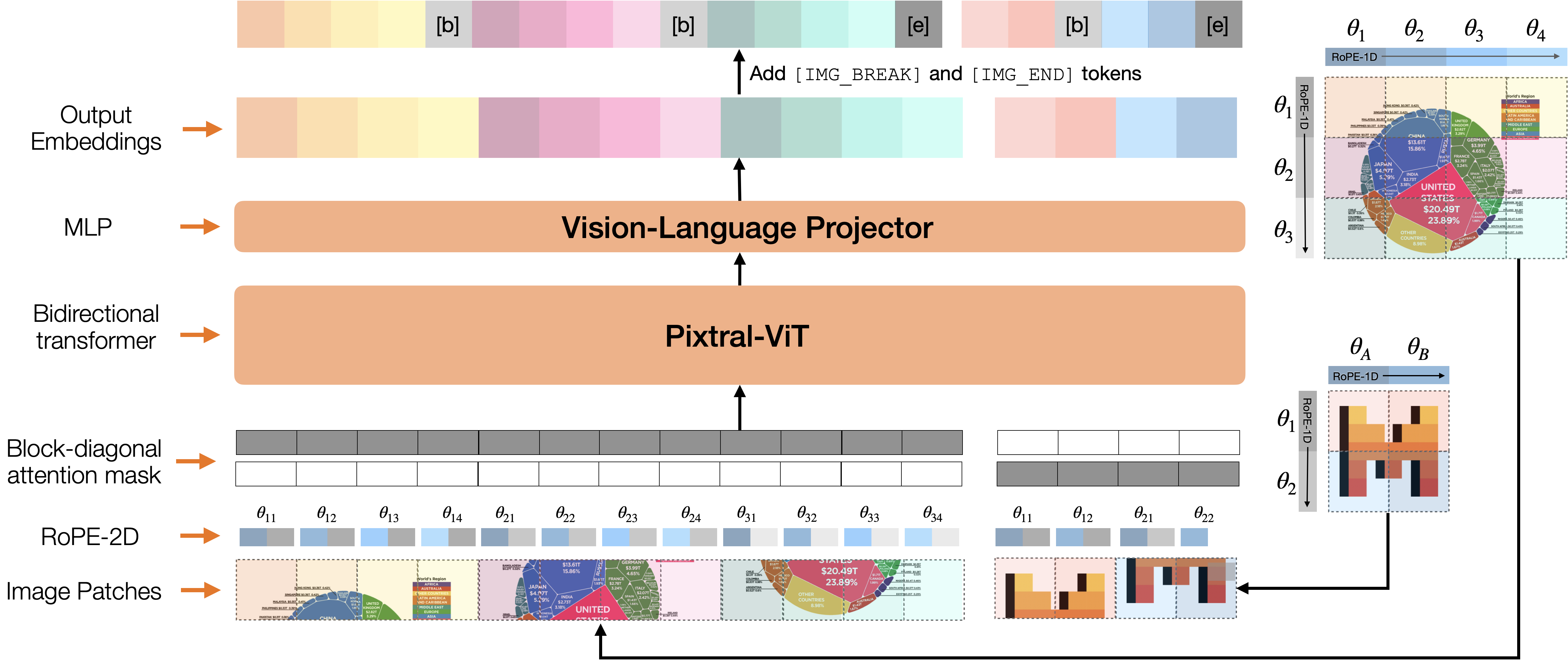}
\caption{\small \textbf{Pixtral Vision Encoder.} Pixtral uses a new vision encoder, which is trained from scratch to natively support variable image sizes and aspect ratios.
Block-diagonal attention masks enable sequence packing for batching, while \textsc{RoPE-2D} encodings facilitate variable image sizes.
Note that the attention mask and position encodings are fed to the vision transformer as additional input, and utilized only in the self-attention layers. 
}
\label{fig:ve}
\end{figure}

In order for Pixtral 12B to ingest images, we train a new vision encoder from scratch, named Pixtral-ViT. 
Here, our goal is to instantiate a simple architecture which is capable of processing images across a wide range of resolutions and aspect ratios.
To do this, we build a 400 million parameter vision transformer~\cite{dosovitskiy2020image} (see Table~\ref{tab:param}) and make four key changes over the standard architectures~\citep{radford2021learning}:

\textbf{Break tokens:} 
In order to assist the model in distinguishing between images with the same number of patches (same area) but different aspect ratios, we include \texttt{[IMAGE BREAK]} tokens between image rows~\cite{bavishi2023fuyu}. 
We further include an \texttt{[IMAGE END]} token at the end of an image sequence. 

\textbf{Gating in FFN:} Instead of standard feedforward layer in the attention block, we use gating in the hidden layer~\citep{shazeer2020glu}. 

\textbf{Sequence packing:} 
In order to efficiently process images within a single batch, we flatten the images along the sequence dimension and concatenate them~\cite{dehghani2024patch}.
We construct a block-diagonal mask to ensure no attention leakage between patches from different images. 

\textbf{RoPE-2D:}
We replace traditional \textit{learned} and \textit{absolute} position embeddings for image patches with \textit{relative}, \textit{rotary} position encodings~\cite{li2022lepard, su2024roformer} in the self-attention layers. 
While learned position embeddings must be interpolated to deal with new image sizes (often at the cost of performance), relative position encodings lend themselves naturally to variable image sizes.

\looseness=-1 Particularly, let $x$ be a $d$-dimensional patch vector (either a key or query feature). 
We denote this feature as $x^{(i,j)}$ when it appears at position $(i,j)$ in the image. Then, the \textsc{RoPE-2D} transform of $x^{(i,j)}$ is expressed as:

\vspace{-5mm}
\begin{footnotesize}
\begin{align*}
    &\textsc{RoPE-2D}\left(x^{(i,j)}, \Theta \right) = M^{(i,j)}_\Theta x^{(i,j)}  \,,  \numberthis \label{eq:rope2d}\\
    \text{where} \qquad M^{(i,j)}_\Theta = & \begin{pmatrix}
\cos i\theta_1 & -\sin i\theta_1 & 0 & 0 & \cdots & 0 & 0 \\
\sin i\theta_1 & \cos i\theta_1 & 0 & 0 & \cdots & 0 & 0 \\
0 & 0 & \cos j\theta_2 & -\sin j\theta_2 & \cdots & 0 & 0 \\
0 & 0 & \sin j\theta_2 & \cos j\theta_2 & \cdots & 0 & 0 \\
\vdots & \vdots & \vdots & \vdots & \ddots & \vdots & \vdots \\
0 & 0 & 0 & 0 & \cdots & \cos j\theta_\frac{d}{2} & -\sin j\theta_\frac{d}{2} \\
0 & 0 & 0 & 0 & \cdots & \sin j\theta_\frac{d}{2} & \cos j\theta_\frac{d}{2}
\end{pmatrix} \,.
\end{align*}
\end{footnotesize}

Here, sub-matrices $M^{(i,j)}_\Theta[k:k+2,k:k+2]$ capture the height position of the feature ($i$) for odd values of dimension $k$, and capture the width position ($j$) for even values of $k$ (1-based indexing). 
Furthermore, $\Theta = [\theta_1 \dots \theta_{d/2}]$ is a vector of frequencies for the various dimensions of $x$, where $\theta_m$ is defined following standard practice for \textsc{RoPE-1D}~\cite{su2024roformer}. 

Critically, our simple implementation of the \textsc{RoPE-2D} transform satisfies the ``relative'' property: that inner products between two vectors are dependent only on their relative difference in height and width position, rather than their absolute position (see more details in Appendix~\ref{sec:appendixrope}).

\begin{figure*}[t]
\vspace{-10pt}
\includegraphics[width=0.99\linewidth]{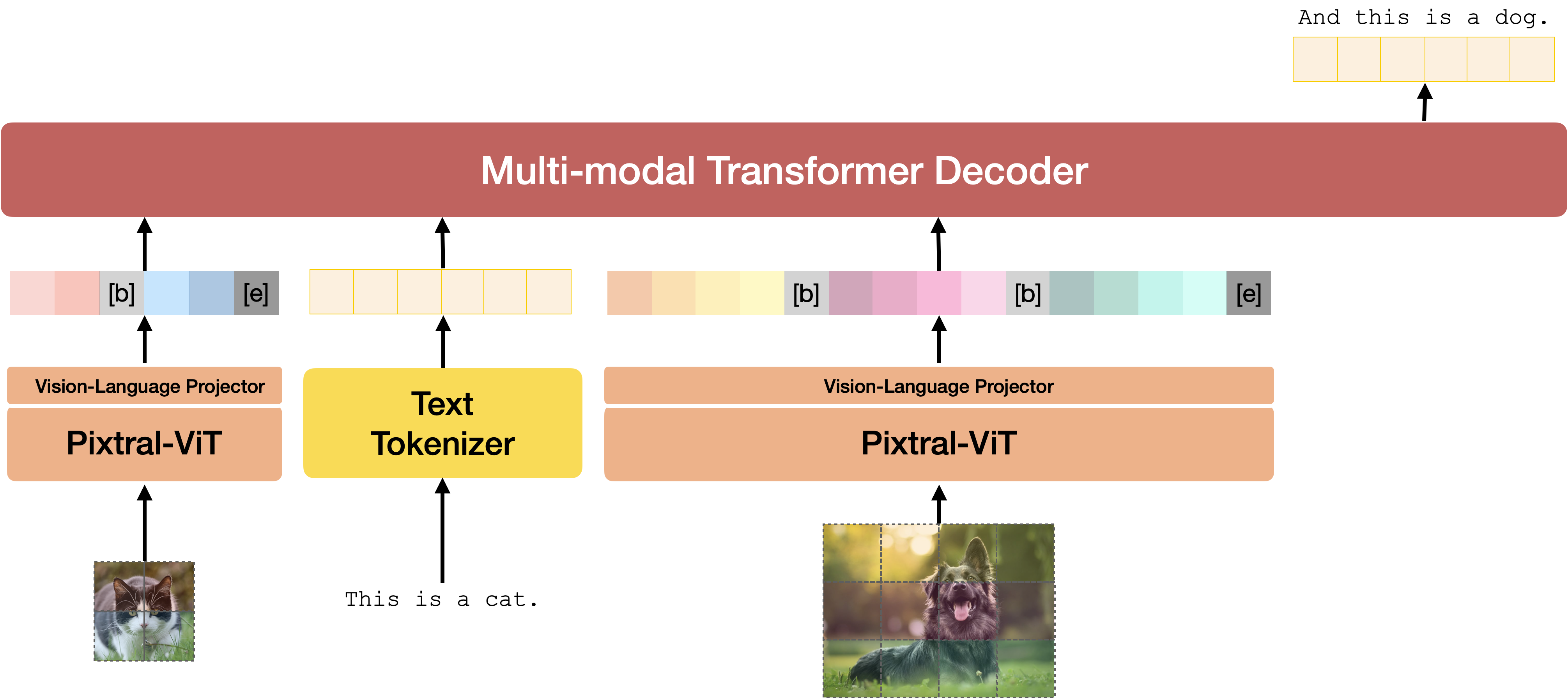}
\caption{
\textbf{Complete Pixtral Architecture.} Pixtral has two components: a \textit{vision encoder}, which tokenizes images, and a \textit{multimodal decoder}, which predicts the next text token given a sequence of text and images. Pixtral can take an arbitrary number of images as input, provided they fit within its 128K context window.
}
\label{fig:full_arch}
\end{figure*}

\textbf{Discussion:} 
Our vision encoder is specifically designed for multimodal modeling. 
Traditional encoders are typically optimized for ImageNet performance at a resolution of, for example, $224 \times 224$ or $336 \times 336$ pixels. 
When incorporated into multimodal language models -- which flexibly perform tasks from standard classification to optical character recognition -- prior works typically break an image into smaller (square) tiles before independently feeding tiles to the vision encoder.
Instead, our vision encoder can naturally adapt to both high and low resolution images at their native aspect ratio, providing substantially improved performance for multi-modal tasks (see Section~\ref{sec:vis_enc_ablations}).

\subsection{Complete architecture}

The Pixtral vision encoder is linked to the multimodal decoder via a two-layer fully connected network. 
This network transforms the output of the vision encoder into the input embedding size required by the decoder via an intermediate hidden layer of the same size, employing the GeLU activation~\cite{hendrycks2016gaussian}. 
The image tokens are treated identically to the text tokens by the multimodal decoder, including \mbox{RoPE-1D}~\cite{su2024roformer} positional encodings for all tokens. 
Particularly, our decoder uses a causal self-attention mechanism, smoothly facilitating capabilities such as multi-image conversations.
The architecture is illustrated in Figure~\ref{fig:full_arch}.

\section{MM-MT-Bench: A benchmark for multi-modal instruction following}

Most existing multimodal benchmarks measure the ability of a model to perform some form of multiple-choice question answering given an input image.
While this is a useful signal for the model's ability to understand the image, it does not capture the extent of the model's utility to a user (for instance as a multimodal assistant or chatbot).
In order to measure this quality, instruction-tuned text-only models are typically evaluated on MT-Bench~\cite{zheng2023judging}, wherein an independent LLM \textit{judge} grades a model's output with respect to a reference answer.
We construct and release a new benchmark named Multimodal MT-Bench (MM-MT-Bench) in a similar vein to the text-only variant, to evaluate the performance of instruction-tuned multimodal models.

\textbf{Design.} MM-MT-Bench contains 92 conversations in total. 
It covers a breadth of practical use cases, covering five categories of images: charts (21), tables (19), PDF pages (24) diagrams (20) and miscellaneous~(8). There are 69 single-turn conversations, 18 conversations with 2 turns, 4 of them with 3 turns and 1 conversation with 4 turns. 
 To evaluate a model, we query the model in parallel over all turns of a conversation, providing reference answers for the past turns as history. Each turn is rated independently by the judge with the entire conversation history provided. 
 The judge is prompted to rate the conversation on a scale of 1 to 10 based on correctness (\textit{i.e.} was the extracted information correct) and completeness (\textit{i.e.} does the model answer cover all the points raised in the reference).
 The evaluation process is illustrated in Figure~\ref{fig:mm_mt_bench}. The judge prompt is provided in Appendix \ref{sec:mm_mt_bench_judge_prompt}. The results shown in Table~\ref{tab:oss_comparisons} show that MM-MT-Bench has a \textbf{0.91 Pearson Correlation Coefficient} with LMSys-Vision ELO ratings.

\begin{figure}[htbp]
  \includegraphics[width=0.99\textwidth]{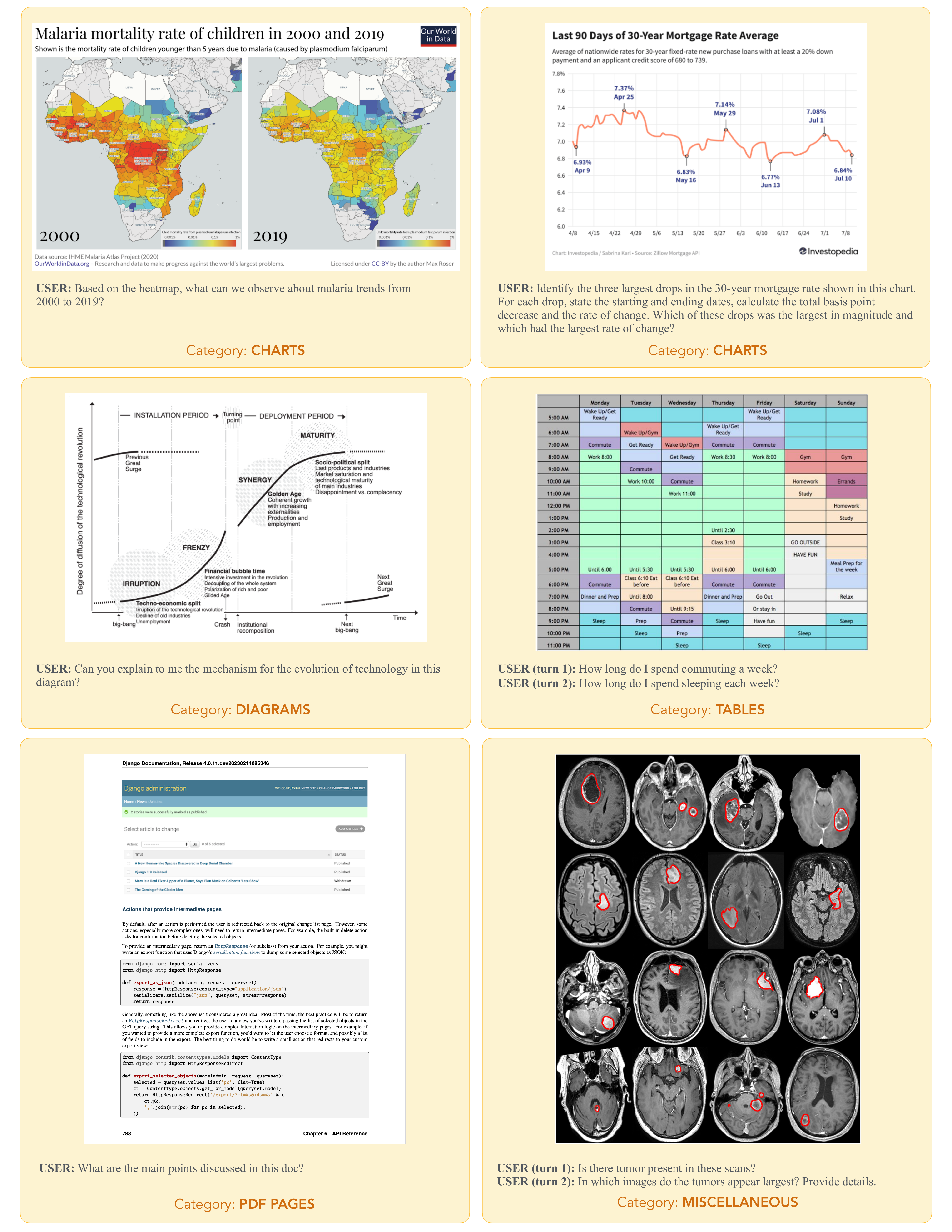}
  \caption{Example images from MM-MT-Bench}
  \label{fig:mm_mt_bench_examples}
\end{figure}

\textbf{Examples.} MM-MT-Bench was designed to mimic real world usage of vision-language models, for extraction, summarization and reasoning over the contents of an image.
Representative images from each category are provided in Figure~\ref{fig:mm_mt_bench_examples} and an example of rated model responses from vision-language models are provided in Figure~\ref{fig:model-responses}.
We manually curated the images, prompts and answers and verified the answers from a second group of labelers.
We ensure that all prompts require reference to the image input to be answered correctly.

\begin{figure*}
    \centering
    \includegraphics[width=0.89\linewidth]{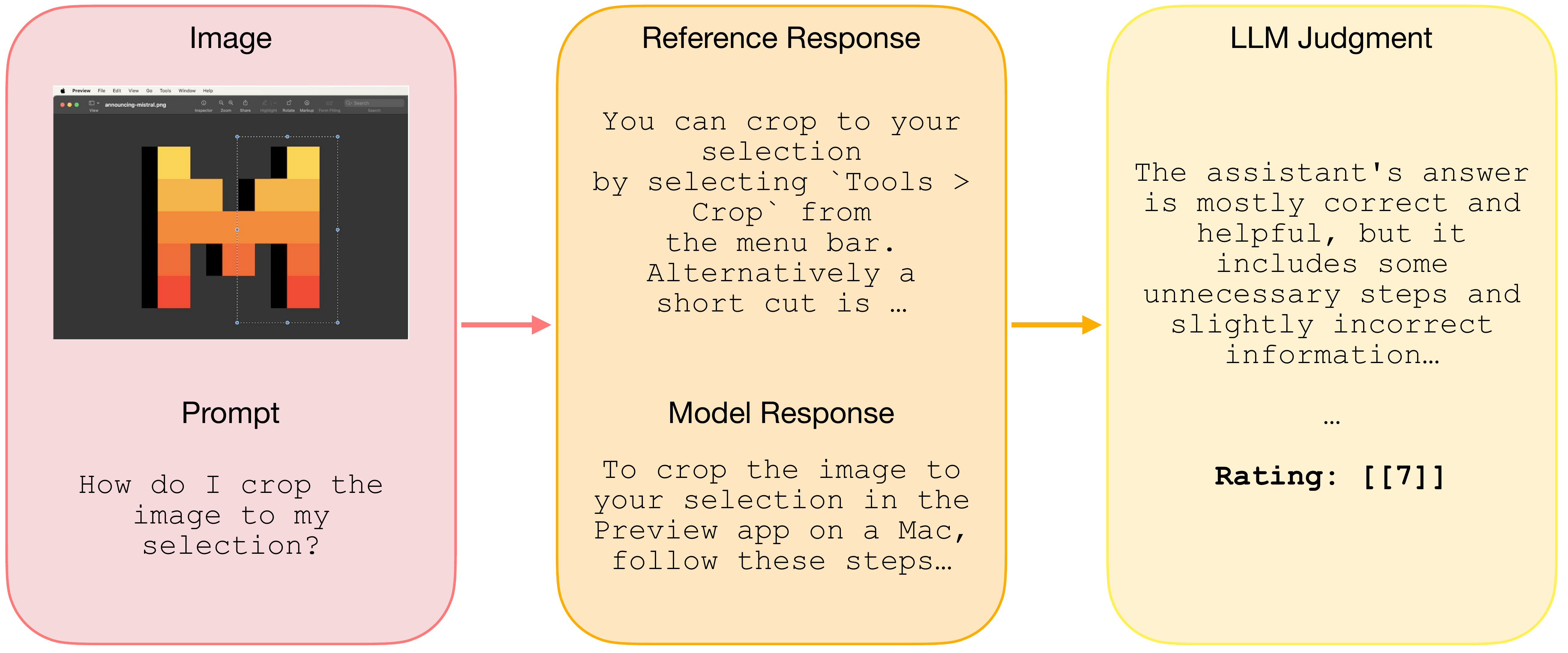}
    \caption{\textbf{MM-MT-Bench:} We open-source a new instruction following benchmark for multimodal models, which correlates highly with LMSys ELO ratings. 
    Given an input image, reference answer and model response, an independent LLM judge is instructed to grade the model's response on a scale of 1 through 10.}
    \label{fig:mm_mt_bench}
\end{figure*}



\definecolor{flexThree}{RGB}{253,224,175} 
\newcommand{\graytext}[1]{\textcolor{gray}{#1}}
\newcommand{\elbowArrow}{%
  \tikz[baseline=-0.5ex]{
    \draw[-, line width=0.5pt] (0,0) -- (0.2,0) -- (0.2,0.4);
    \draw[->, line width=0.5pt] (0.2,0.4) -- (0.4,0.4);
  }%
}

\begin{table}
\small
\renewcommand{\arraystretch}{1.2}
\resizebox{\linewidth}{!}{%
\begin{tabular}{@{}l*{7}{c}@{}}
\toprule
& \textbf{Mathvista} & \textbf{MMMU} & \textbf{ChartQA} & \textbf{DocVQA} & \textbf{VQAv2} & \textbf{MM-MT-Bench} & \textbf{LMSys-Vision} \\[-2pt]
& \textcolor{gray}{\scriptsize CoT} & \textcolor{gray}{\scriptsize CoT} & \textcolor{gray}{\scriptsize CoT} & \textcolor{gray}{\scriptsize ANLS} & \textcolor{gray}{\scriptsize VQA Match} & \textcolor{gray}{\scriptsize GPT-4o Judge} & \textcolor{gray}{\scriptsize (Oct '24)}  \\
\midrule
\rowcolor{flexThree}
Pixtral 12B & \textbf{58.3} & \textbf{52.0} & \textbf{81.8} & 90.7 & \textbf{78.6} & \textbf{6.05} & 1076 \\
Qwen-2-VL 7B~\cite{wang2024qwen2vlenhancingvisionlanguagemodels} & 53.7 & 48.1 & 41.2 & \textbf{94.5} & 75.9 & 5.45 & 1040 \\
\textcolor{gray}{$\quad$→ w/ Flexible Parsing} & \textcolor{gray}{55.2} & \textcolor{gray}{48.7} & \textcolor{gray}{77.5} & \textcolor{gray}{--} & \textcolor{gray}{--} & \textcolor{gray}{--} & \textcolor{gray}{--} \\
Llama-3.2 11B~\cite{dubey2024llama3} & 24.3 & 23.0 & 14.8 & 91.1 & 67.1 & 4.79 & 1032 \\
\textcolor{gray}{$\quad$→ w/ Flexible Parsing} & \textcolor{gray}{47.9} & \textcolor{gray}{45.3} & \textcolor{gray}{78.5} & \textcolor{gray}{--} & \textcolor{gray}{--} & \textcolor{gray}{--} & \textcolor{gray}{--} \\
Molmo-D 7B~\cite{deitke2024molmo} & 12.3 & 24.3 & 27.0 & 72.2 & 57.1 & 3.72 & -- \\
LLaVA-OneVision 7B~\cite{li2024llava} & 36.1 & 45.1 & 67.2 & 90.5 & 78.4 & 4.12 & -- \\
\midrule
Claude-3 Haiku~\cite{claude3} & 44.8 & 50.4 & 69.6 & 74.6 & 68.4 & 5.46 & 1000 \\
Gemini-1.5-Flash 8B\textcolor{gray}{\scriptsize (0827)}~\cite{reid2024gemini} & 56.9 & 50.7 & 78.0 & 79.5 & 65.5 & 5.93 & \textbf{1111} \\
\midrule
\graytext{Molmo 72B~\cite{deitke2024molmo}} & \graytext{52.2} & \graytext{52.7} & \graytext{75.6} & \graytext{86.5} & \graytext{75.2} & \graytext{3.51} & \graytext{--} \\
\graytext{LLaVA-OneVision 72B~\cite{li2024llava}} & \graytext{57.2} & \graytext{54.4} & \graytext{66.9} & \graytext{91.6} & \graytext{83.8} & \graytext{4.95} & \graytext{992} \\
\graytext{Qwen-2-VL 72B~\cite{wang2024qwen2vlenhancingvisionlanguagemodels}} & \graytext{68.2} & \graytext{60.3} & \graytext{66.6} & \graytext{96.3} & \graytext{81.6} & \graytext{6.59} & \graytext{1104} \\
\graytext{Llama-3.2 90B~\cite{dubey2024llama3}} & \graytext{49.1} & \graytext{53.7} & \graytext{33.8} & \graytext{85.7} & \graytext{67.0} & \graytext{5.50} & \graytext{1071} \\
\midrule
\graytext{GPT-4o~\textcolor{gray}{\scriptsize (0513)}~\cite{openai2023gpt}} & \graytext{64.6} & \graytext{68.6} & \graytext{85.1} & \graytext{88.9} & \graytext{77.8} & \graytext{7.72} & \graytext{1208} \\
\graytext{Claude-3.5 Sonnet~\cite{claude3}} & \graytext{64.4} & \graytext{68.0} & \graytext{87.6} & \graytext{90.3} & \graytext{70.7} & \graytext{7.50} & \graytext{1189} \\
\bottomrule
\end{tabular}%
}
\vspace{1mm}
\caption{\textbf{Multimodal Benchmarks.}
Pixtral substantially outperforms open models of a similar size, as well as several closed-source models.
We re-evaluate all models with the same prompt and evaluation metric (see Section~\ref{sec:prompt_selection}).
For transparent comparison against Qwen2-VL 7B~\cite{wang2024qwen2vlenhancingvisionlanguagemodels} and Llama-3.2 11B~\cite{dubey2024llama3}, we additionally report their performance under relaxed evaluation constraints in \textcolor{gray}{(gray)} (see Section~\ref{sec:parsing_ablation}).
To further investigate the gap with reported figures for some open-source models, we provide analysis in Section~\ref{sec_app:reproducing_reported}.
}
\label{tab:oss_comparisons}
\end{table}

\begin{table}
\small
\renewcommand{\arraystretch}{1.2}
\centering
\resizebox{0.7\linewidth}{!}{%
\begin{tabular}{@{}l*{4}{c}@{}}
\toprule
& \textbf{MT-Bench} & \textbf{MMLU} & \textbf{Math} & \textbf{HumanEval} \\
& & \textcolor{gray}{\scriptsize 5-shot} & \textcolor{gray}{\scriptsize Maj@1} & \textcolor{gray}{\scriptsize Pass@1} \\
\midrule
\rowcolor{flexThree}
Pixtral 12B & \textbf{7.68} & \textbf{69.2} & 48.1 & \textbf{72.0} \\
\midrule
LLaVA-OneVision 7B~\cite{li2024llava} & 6.94 & 67.9 & 38.6 & 65.9 \\
Molmo-D 7B~\cite{deitke2024molmo} & 4.53 & 61.2 & 10.2 & 3.7 \\
Qwen-2-VL 7B~\cite{wang2024qwen2vlenhancingvisionlanguagemodels} & 6.41 & 68.5 & 27.9 & 62.2 \\
Llama-3.2 11B~\cite{dubey2024llama3} & 7.51 & 68.5 & \textbf{48.3} & 62.8 \\
\bottomrule
\end{tabular}%
}
\vspace{1mm}
\caption{
\textbf{Language benchmarks.}
Pixtral 12B consistently outperforms open-source models of a comparable size on text-only benchmarks, making it a drop-in multimodal replacement for existing text-only deployments.
}
\label{tab:closed_source_comparisons}
\end{table}

\section{Results}

In this section, we provide evaluations of Pixtral 12B against closed and open-source models across a range of model sizes, re-evaluating all models through the same evaluation harness. 
Particularly, for each dataset, we design the prompt such that we can reproduce the results of leading multimodal models (GPT-4o~\cite{openai2023gpt} and Claude-3.5 Sonnet~\cite{claude3}).
These prompts are `Explicit' and fully specify the output format (see Section~\ref{sec:prompt_selection}), allowing models which follow the prompt instructions to be marked accurately at test-time.
All models were evaluated with the same prompts, which are specified in Appendix~\ref{sec:appendixA}.
We provide additional analysis on re-evaluating models under various prompts and metrics in Sections~\ref{sec:prompt_selection} and ~\ref{sec:parsing_ablation}, as well as in Appendices~\ref{sec_app:average_results} and ~\ref{sec_app:reproducing_reported}.

\subsection{Main Results}
\label{sec:main_results}
\textbf{Multimodal performance:}
Table~\ref{tab:oss_comparisons} shows that Pixtral substantially outperforms all open models around its scale on multimodal benchmarks, as well as closed source models such as Claude-3 Haiku~\cite{claude3} and Gemini-1.5 Flash 8B~\cite{reid2024gemini}.
Particularly, Pixtral outperforms all models of comparable size on MM-MT-Bench, which targets real world use cases, a finding corroborated by strong performance on LMSys Vision Arena.
On this public leaderboard, Pixtral 12B approaches the performance of the largest open-weights models, such as Qwen2-VL 72B~\cite{wang2024qwen2vlenhancingvisionlanguagemodels} and Llama-3.2 90B~\cite{dubey2024llama3}. 

We highlight that, with our `Explicit' prompts, the performance of some open-source models is substantially lower than their reported figures. 
For the closest open-source models -- Qwen2-VL 7B~\cite{wang2024qwen2vlenhancingvisionlanguagemodels} and Llama-3.2 11B~\cite{dubey2024llama3} -- this is mainly due to models not following instructions on answer formatting (\textit{e.g.} generating \texttt{"The answer is 6."} instead of \texttt{"Final answer: 6"}).
For transparent comparison against these models, we further report their evaluations using relaxed metrics, with more flexible parsing, in \textcolor{gray}{gray} (see Section~\ref{sec:parsing_ablation}).
We analyze the performance of these models under various prompts in Appendix~\ref{sec_app:average_results}.
In Appendix~\ref{sec_app:reproducing_reported}, we customize the evaluation to each model in turn, describing the changes required to bridge the gaps to reported performance.

\textbf{Language performance:}
Table~\ref{tab:closed_source_comparisons} evaluates Pixtral 12B against open-source models of comparable size on common text-only benchmarks (again, with common prompting and evaluation protocols). 
Pixtral does not compromise text understanding in pursuit of multimodal capabilities, making it a suitable drop-in replacement for both text and vision tasks.

\subsection{Prompt selection}
\label{sec:prompt_selection}

Here we discuss our methodology for designing the evaluation prompts.
In our evaluation harness, we choose prompts which allow for reproduction of the reported results of leading closed-source models: GPT-4o~\cite{openai2023gpt} and Claude-3.5-Sonnet~\cite{claude3}.
These prompts are provided in Appendix~\ref{sec:appendixA}, and we report results averaged over 10 prompts in Appendix~\ref{sec_app:average_results}.

\begin{figure}
    \centering
    \includegraphics[width=\linewidth]{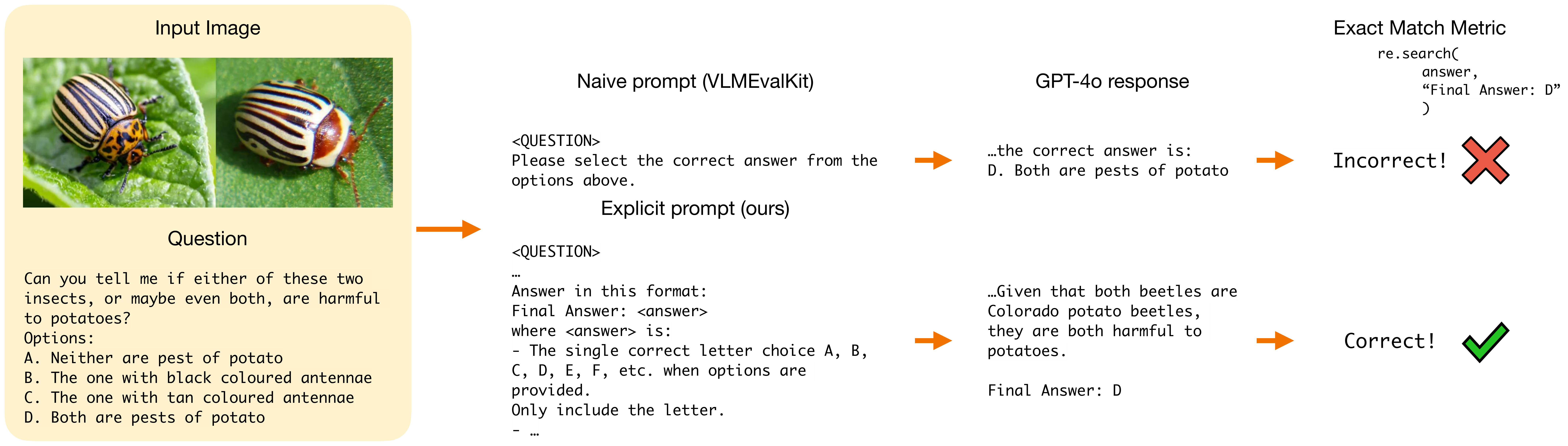}
    \caption{
    \textbf{Effect of `Naive' vs. `Explicit' prompts on leading models.}
    Leading models benefit greatly from `Explicit' prompts which provide details about the output format.
    This makes sense, as otherwise substantively correct responses are marked as incorrect during evaluation (top row, right).
    }
    \label{fig:prompt_selection}
\end{figure}

We find that commonly used prompts do not properly specify the output format.
For instance, for a multiple choice question, we find open-source prompts include vague instructions like \texttt{"Select the correct answer from the options above"}. 
In this case, it is impossible for models to know whether answers should be presented as an index (\texttt{"Option A"}, \texttt{"Option B"} \textit{etc.}) or with a natural language response.
Models are then penalized for incorrect formatting.
As such, leading models require prompts which \textit{explicitly} specify the required output format. 
We illustrate this with a real example from MMMU in Figure~\ref{fig:prompt_selection}.

In Table~\ref{tab:prompt_albations}, we demonstrate that our `Explicit' prompts substantially improve the performance of leading models over `Naive' prompts.
We also note that in a number of cases, the performance of smaller models \textit{reduces} with the Explicit prompt format, perhaps due to a discrepancy with the prompt-style in the training set of these benchmarks.
Pixtral 12B generally performs better with Explicit prompts, with only a minor regression on ChartQA.

\definecolor{flexOne}{RGB}{199,232,229}   
\definecolor{flexTwo}{RGB}{239,245,214}   
\definecolor{flexThree}{RGB}{253,224,175} 
\definecolor{lightgreen}{RGB}{200,255,200}

\begin{table}
\centering
\small
\begin{tabular}{@{}lcccccc@{}}
\toprule
& \multicolumn{2}{c}{\cellcolor{flexOne}\textbf{VQAv2}} & \multicolumn{2}{c}{\cellcolor{flexTwo}\textbf{ChartQA}} & \multicolumn{2}{c}{\cellcolor{flexThree}\textbf{MMMU}} \\
\cmidrule(lr){2-3} \cmidrule(lr){4-5} \cmidrule(lr){6-7}
Prompt $\longrightarrow$ & Naive & Explicit & Naive & Explicit & Naive & Explicit \\
\midrule
GPT-4o~\textcolor{gray}{\scriptsize (0513)}~\cite{openai2023gpt} & 64.2 & \cellcolor{lightgreen}77.8 & 58.0 & \cellcolor{lightgreen}85.1 & 55.0 & \cellcolor{lightgreen}68.6 \\
Sonnet-3.5~\cite{claude3} & 50.2 & \cellcolor{lightgreen}70.7 & 39.6 & \cellcolor{lightgreen}87.6 & 48.6 & \cellcolor{lightgreen}68.0 \\
\midrule
Qwen-2-VL 7B~\cite{wang2024qwen2vlenhancingvisionlanguagemodels} & \cellcolor{lightgreen}82.1 & 75.9 & \cellcolor{lightgreen}83.4 & 41.2 & 46.7 & \cellcolor{lightgreen}48.1 \\
Llama-3.2 11B~\cite{touvron2023llama} & 29.5 & \cellcolor{lightgreen}67.1 & 0.0 & \cellcolor{lightgreen}14.8 & 20.7 & \cellcolor{lightgreen}23.0 \\
Llama-3.2 90B~\cite{touvron2023llama} & 52.6 & \cellcolor{lightgreen}67.0 & 3.9 & \cellcolor{lightgreen}33.8 & 27.0 & \cellcolor{lightgreen}53.7 \\
Pixtral 12B & 78.9 & \cellcolor{lightgreen}78.6 & \cellcolor{lightgreen}84.3 & 81.8 & 45.8 & \cellcolor{lightgreen}52.0 \\
\bottomrule
\end{tabular}
\vspace{1mm}
\caption{\textbf{Prompt ablations.}
Leading models require prompts which explicitly specify the output format to perform well. 
Pixtral 12B performs well with both `Explicit' and `Naive' prompts, with only a minor regression on ChartQA.
}
\label{tab:prompt_albations}
\end{table}

\subsection{Sensitivity to evaluation metrics}
\label{sec:parsing_ablation}

In Section~\ref{sec:prompt_selection}, we discuss the importance of prompts which properly specify the output format.
However, during evaluations, we find that even with Explicit prompts, many models still provide outputs in various formats, which are then penalized by metrics which require responses to match the reference answers exactly.

To investigate this, we take models' generations and evaluate them under progressively looser parsing constraints.
For instance, if the correct answer is \texttt{"6"}, flexible metrics do not penalize answers such as \texttt{"6.0"} or \texttt{"The answer is 6"}. 
We provide the details of these parsing settings in Appendix~\ref{sec_app:parsing_ablations}, but here note that `Flexible Level 3' marks a response as correct if the reference answer occurs \textit{anywhere} in the generation. 
This is an overly generous metric which is included only to illustrate an upper bound, as it permits answers like \texttt{"6000"} for a reference answer of \texttt{"6"}.

We provide the results of our analysis in Table~\ref{tab:parsing_ablations}.
We find that the performance of some models dramatically improves with more flexible parsing metrics, indicating that the lower scores can be attributed to the inability of models to properly follow prompt instructions.
We further note that Pixtral 12B benefits very little from flexible parsing (substantiating its ability to follow instructions), and furthermore can generally outperform other models even after flexible metrics are used.

\definecolor{flexOne}{RGB}{199,232,229}   
\definecolor{flexTwo}{RGB}{239,245,214}   
\definecolor{flexThree}{RGB}{253,224,175} 
\definecolor{stdDevColor}{gray}{0.5}      
\newcommand{\stdDevSize}{\scriptsize}
\newcommand{\stdDev}[1]{{\stdDevSize\textcolor{stdDevColor}{($\pm$#1)}}}
\begin{table}
\centering
\small
\begin{tabular}{@{}lcccc@{}}
\toprule
 & Llama-3.2 11B~\cite{touvron2023llama} & Llama-3.2 90B~\cite{touvron2023llama} & Qwen2-VL 7B~\cite{wang2024qwen2vlenhancingvisionlanguagemodels} & Pixtral 12B \\
\midrule
\multicolumn{5}{@{}l}{\textbf{Mathvista}} \\
\midrule
Baseline & 24.3 & 49.1 & 53.7 & \textbf{58.3} \\
\rowcolor{flexOne}
Flexible level 1 & 25.9 & 50.3 & 54.3 & \textbf{58.3} \\
\rowcolor{flexTwo}
Flexible level 2 & 40.2 & 54.7 & 54.3 & \textbf{58.3} \\
\rowcolor{flexThree}
Flexible level 3 & 47.9 & 57.3 & 55.2 & \textbf{58.5} \\
\midrule
\multicolumn{5}{@{}l}{\textbf{MMMU}} \\
\midrule
Baseline & 23.0 & \textbf{53.7} & 48.1 & 52.0 \\
\rowcolor{flexOne}
Flexible level 1 & 23.4 & \textbf{53.7} & 48.1 & 52.0 \\
\rowcolor{flexTwo}
Flexible level 2 & 41.0 & \textbf{55.7} & 48.1 & 52.0 \\
\rowcolor{flexThree}
Flexible level 3 & 45.3 & \textbf{56.7} & 48.7 & 52.0 \\
\midrule
\multicolumn{5}{@{}l}{\textbf{ChartQA}} \\
\midrule
Baseline & 14.8 & 33.8 & 41.2 & \textbf{81.8} \\
\rowcolor{flexOne}
Flexible level 1 & 20.4 & 33.9 & 73.8 & \textbf{81.9} \\
\rowcolor{flexTwo}
Flexible level 2 & 29.9 & 35.6 & 73.8 & \textbf{81.9} \\
\rowcolor{flexThree}
Flexible level 3 & 78.5 & 79.1 & 77.5 & \textbf{82.0} \\
\bottomrule
\end{tabular}
\vspace{1mm}
\caption{\textbf{Flexible parsing ablations.} 
We evaluate models under progressively looser parsing constraints (see Appendix~\ref{sec_app:parsing_ablations} for details).
Under loose parsing constraints, the performance of some models dramatically improves.
Pixtral 12B performance is stable under all parsing conditions, and continues to lead even when flexible parsing is accounted for.
`Flexible Level 3' is included for illustration only, as it allows some incorrect answers to be marked as correct.
}
\label{tab:parsing_ablations}
\end{table}

\subsection{Vision Encoder Ablations}
\label{sec:vis_enc_ablations}

\begin{figure}
    \centering
    \includegraphics[width=0.6\linewidth]{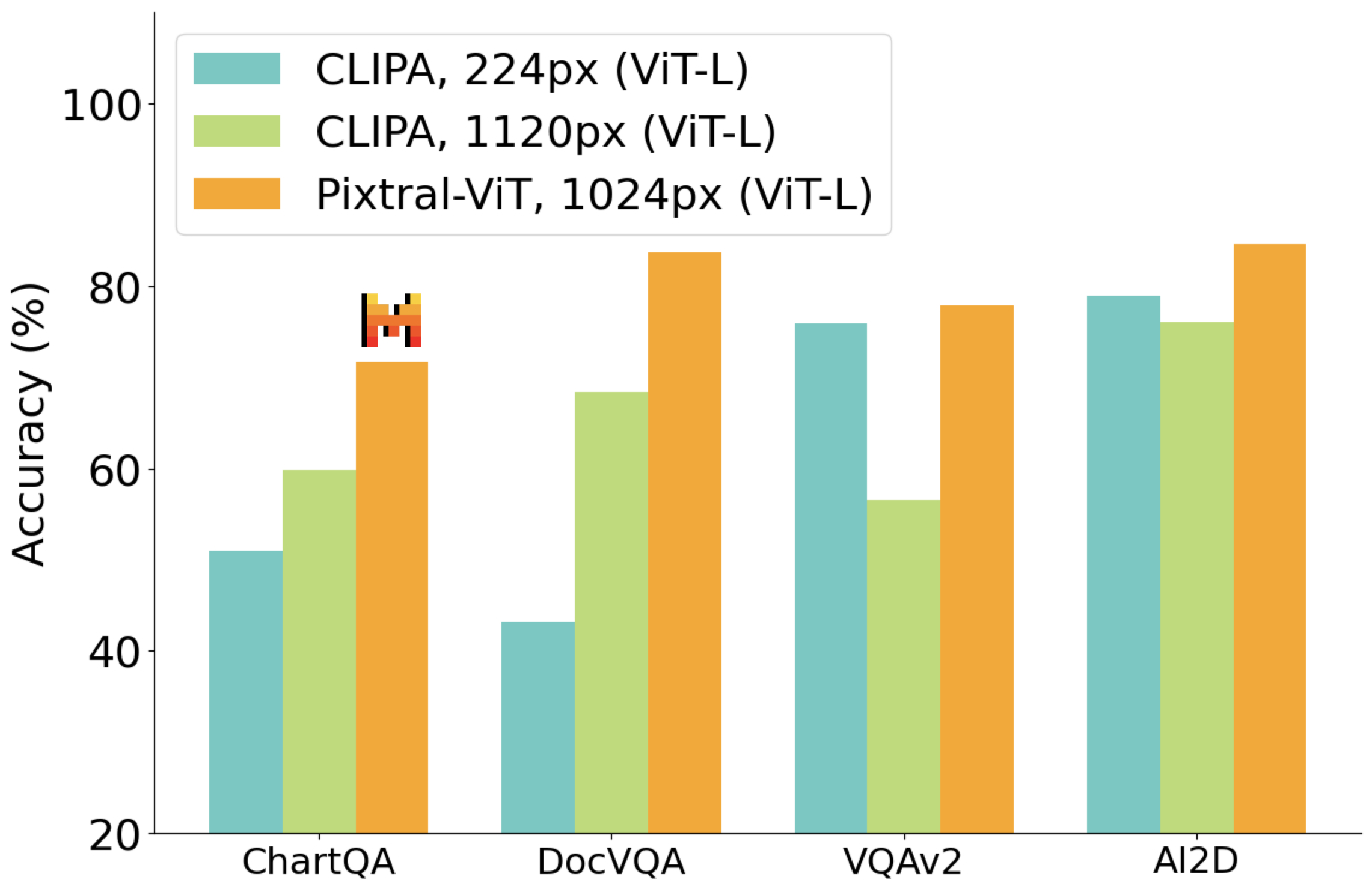}
    \caption{\textbf{Vision encoder ablations:} When leveraged for visual instruction tuning, our encoder substantially outperforms a strong CLIPA~\cite{li2023clipa} baseline for tasks requiring fine-grained document understanding, while maintaining parity for natural images.}
    \label{fig:vis_encoder_ablations}
\end{figure}

In order to verify the design choices for our vision encoder, we conduct small-scale ablations with Visual Instruction Tuning~\cite{liu2024visual}.
We conduct short-horizon multimodal instruction-tuning runs, both with our vision encoder (Pixtral-ViT), as well as a CLIPA~\cite{li2023clipa} backbone as a baseline. For both vision encoders, we use Mistral-Nemo 12B-Instruct~\cite{mistralnemo12b} to initialize the multimodal decoder.

Like many open-source vision encoders, CLIPA is trained at a fixed resolution of $224\times224$ pixels.  
In order to upscale the resolution in vision-language models, existing methods~\cite{liu2024improved} construct several tiled crops from the image, and pass each crop independently through the vision encoder at its pretraining resolution.
We conduct two ablations with CLIPA: (a) we resize the entire image to $224\times224$; (b) we construct $25$ crops of the input image, for a total resolution of $1120\times1120$. 
These models are also evaluated at $224$ pixels and $1120$ pixels respectively, while our flexible encoder is evaluated at variable image resolutions, with a maximum resolution of $1024$ pixels. 

In Figure~\ref{fig:vis_encoder_ablations}, we find that our model substantially outperforms CLIPA in settings which require fine-grained understanding, such as chart and document understanding, while matching its performance on natural language benchmarks such as VQAv2.

\section{Qualitative examples}
\begin{figure*}
\centering
\includegraphics[width=0.5\linewidth,height=\textheight,keepaspectratio]{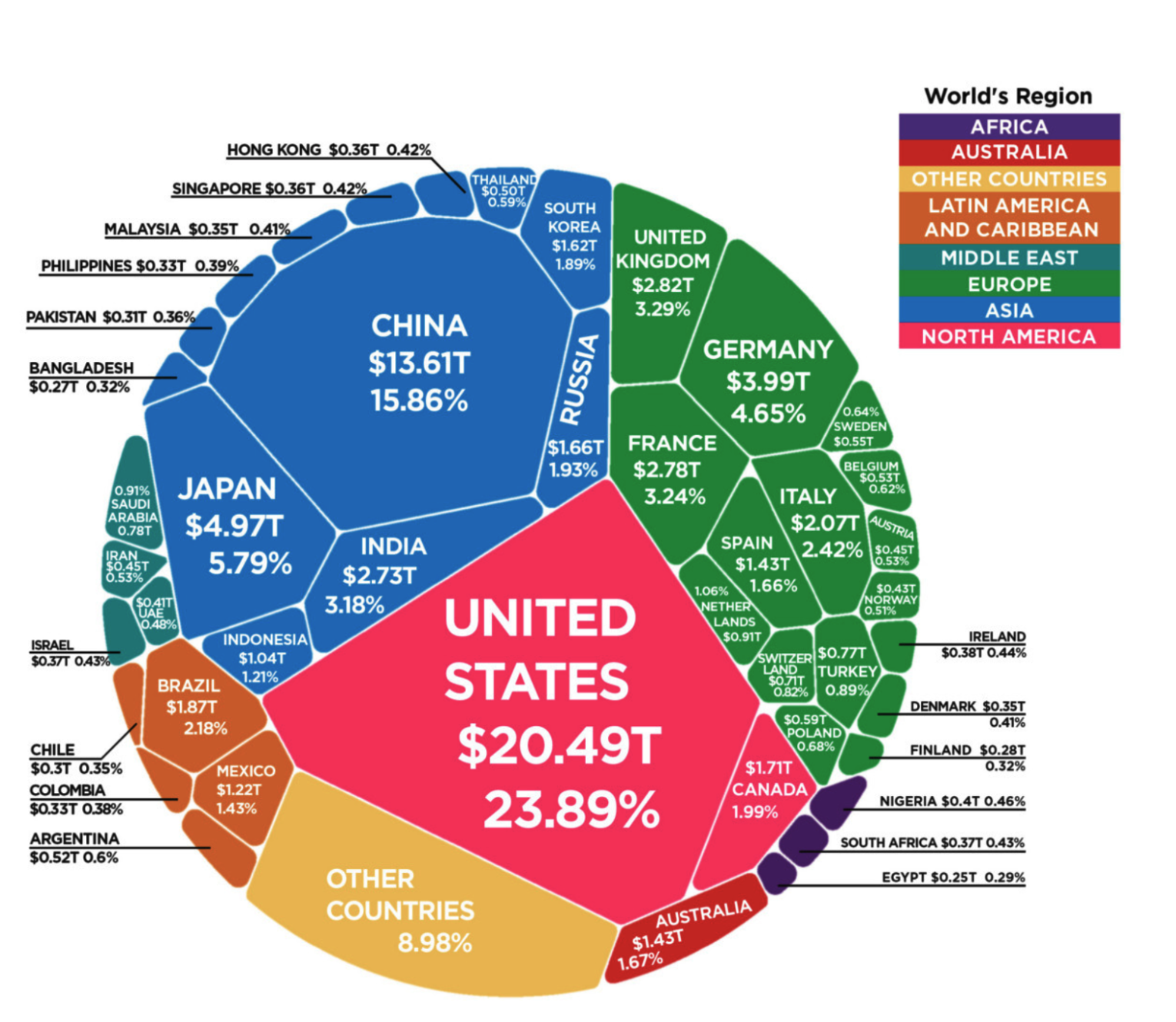}
\includegraphics[width=0.7\linewidth,height=\textheight,keepaspectratio]{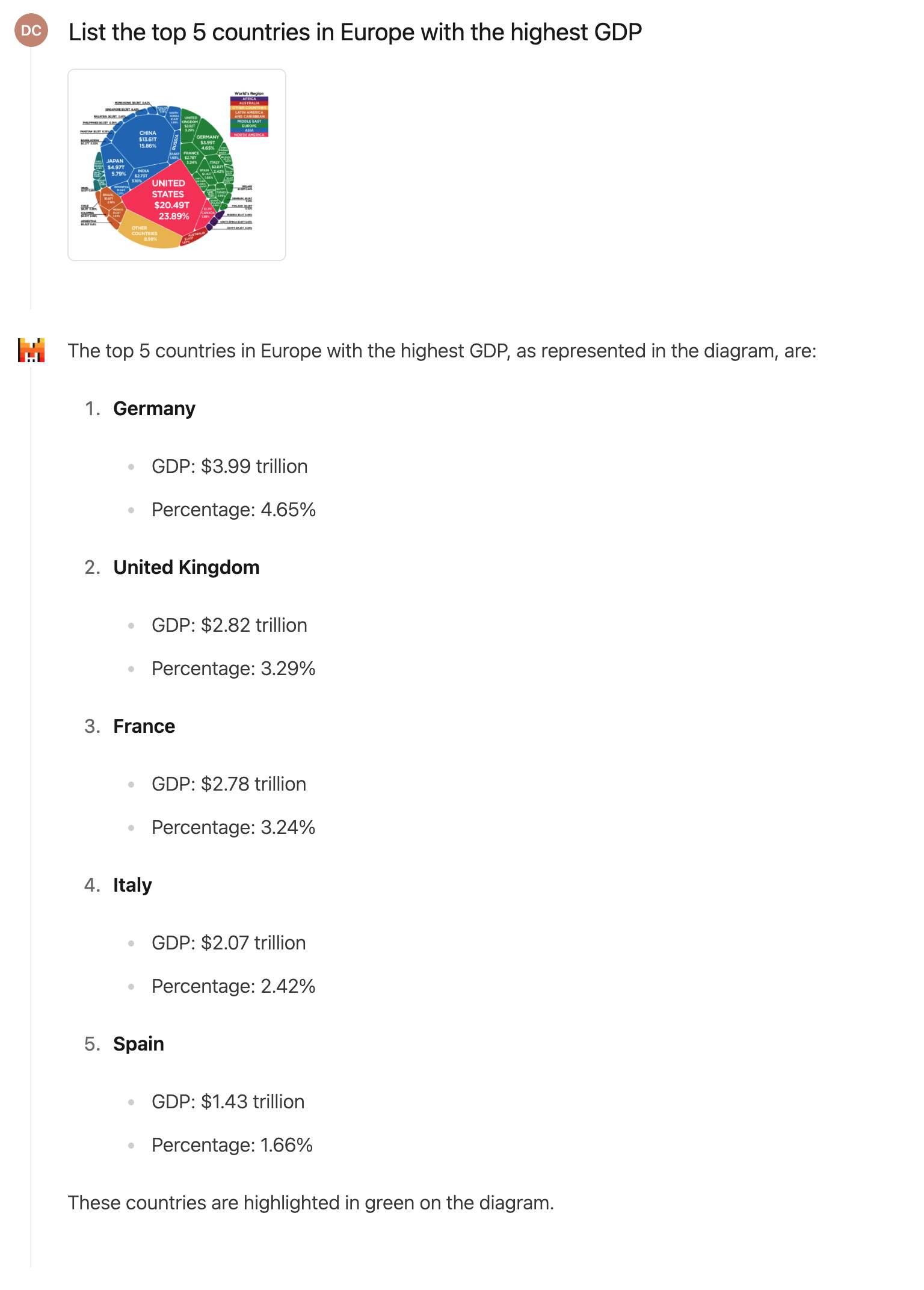}
\caption{\small \textbf{Reasoning over complex figures.} An example showcasing Pixtral's capabilities to understand and reason over complex figures. Pixtral correctly identifies that the green boxes represent the European countries and then reads and sorts the GDP of all the European countries to list the top 5 with accurate GDP numbers.}
\label{fig:eg_gdp}
\end{figure*}

\begin{figure*}
\centering
\includegraphics[width=0.99\linewidth,height=\textheight,keepaspectratio]{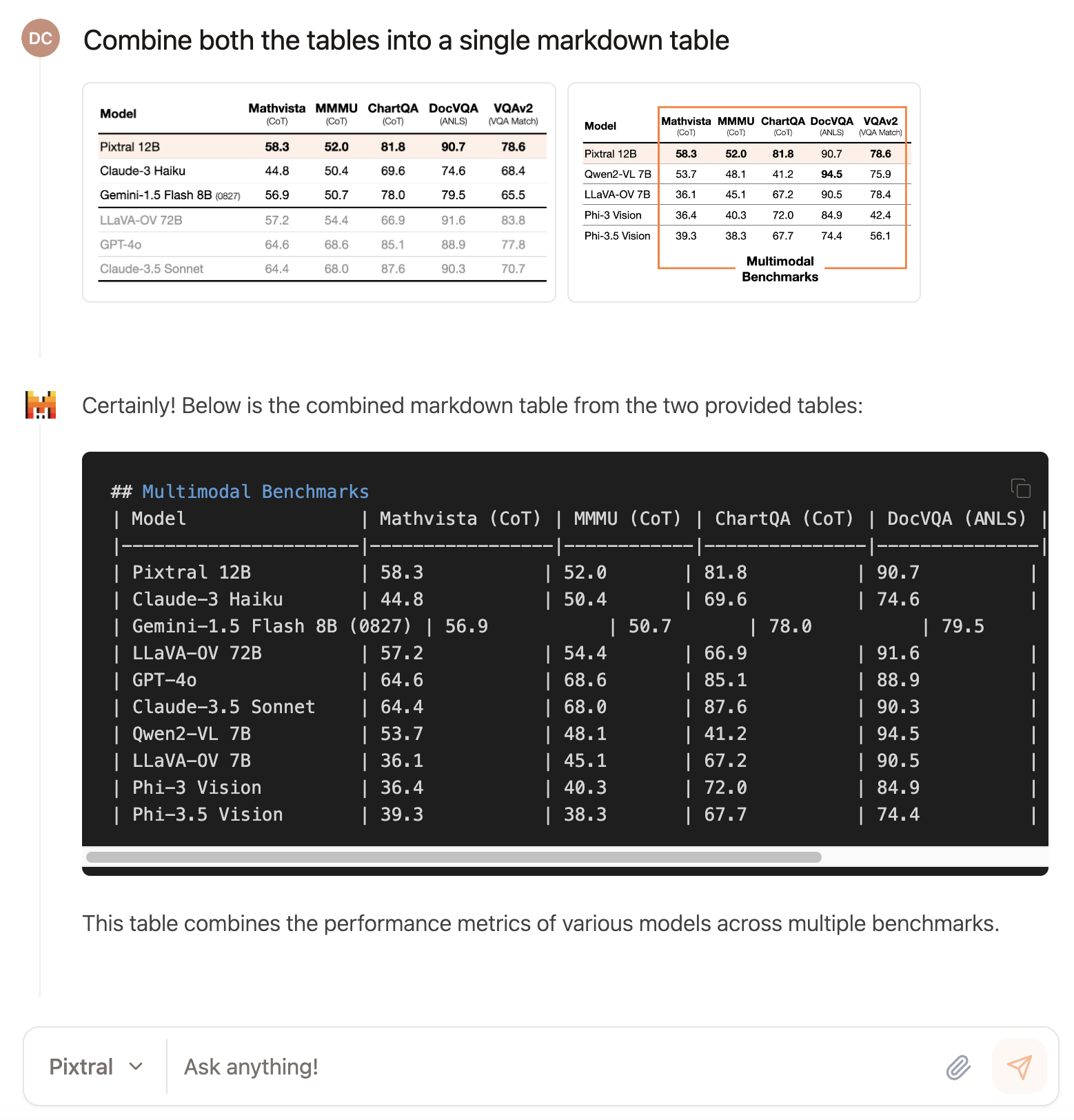}
\caption{\small \textbf{Multi-image instruction following.} Pixtral can process arbitrary number of images in its context window. The example shows that Pixtral can successfully combine the information from both images into a single markdown table.}
\label{fig:eg_multi_image}
\end{figure*}

\begin{figure*}
\centering
\includegraphics[width=0.99\linewidth,height=\textheight,keepaspectratio]{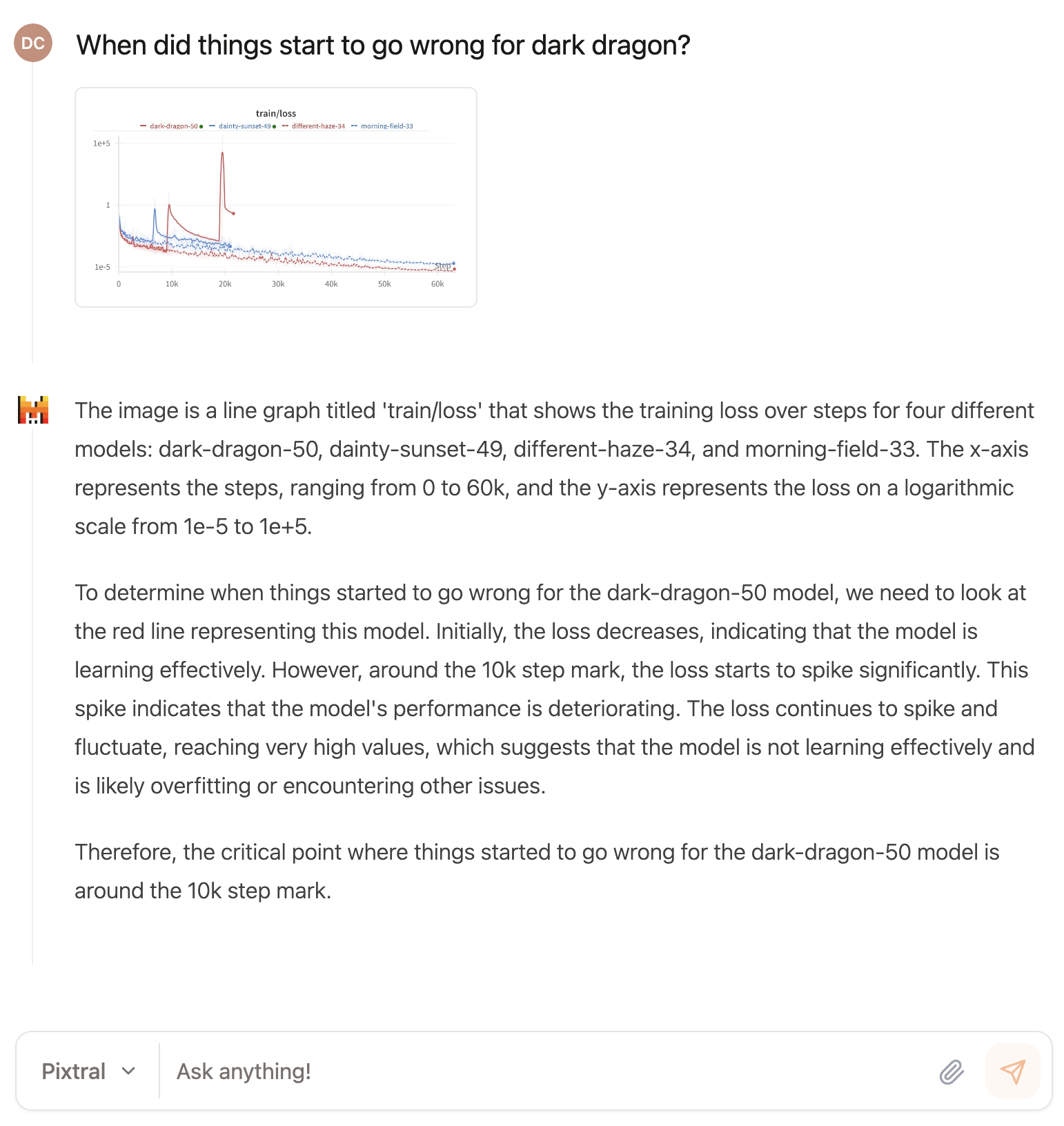}
\caption{\small \textbf{Chart Understanding and Analysis.} Pixtral demonstrates the capability to interpret and analyze intricate charts with high accuracy. In this instance, Pixtral correctly identifies that "dark-dragon" corresponds to the red line. Furthermore, it recognizes that the training loss is expected to decrease smoothly and notes that the training run became unstable around the 10K step mark due to a significant spike in loss.}
\label{fig:eg_chart}
\end{figure*}

\begin{figure*}
\centering
\includegraphics[width=0.9\linewidth,height=\textheight,keepaspectratio]{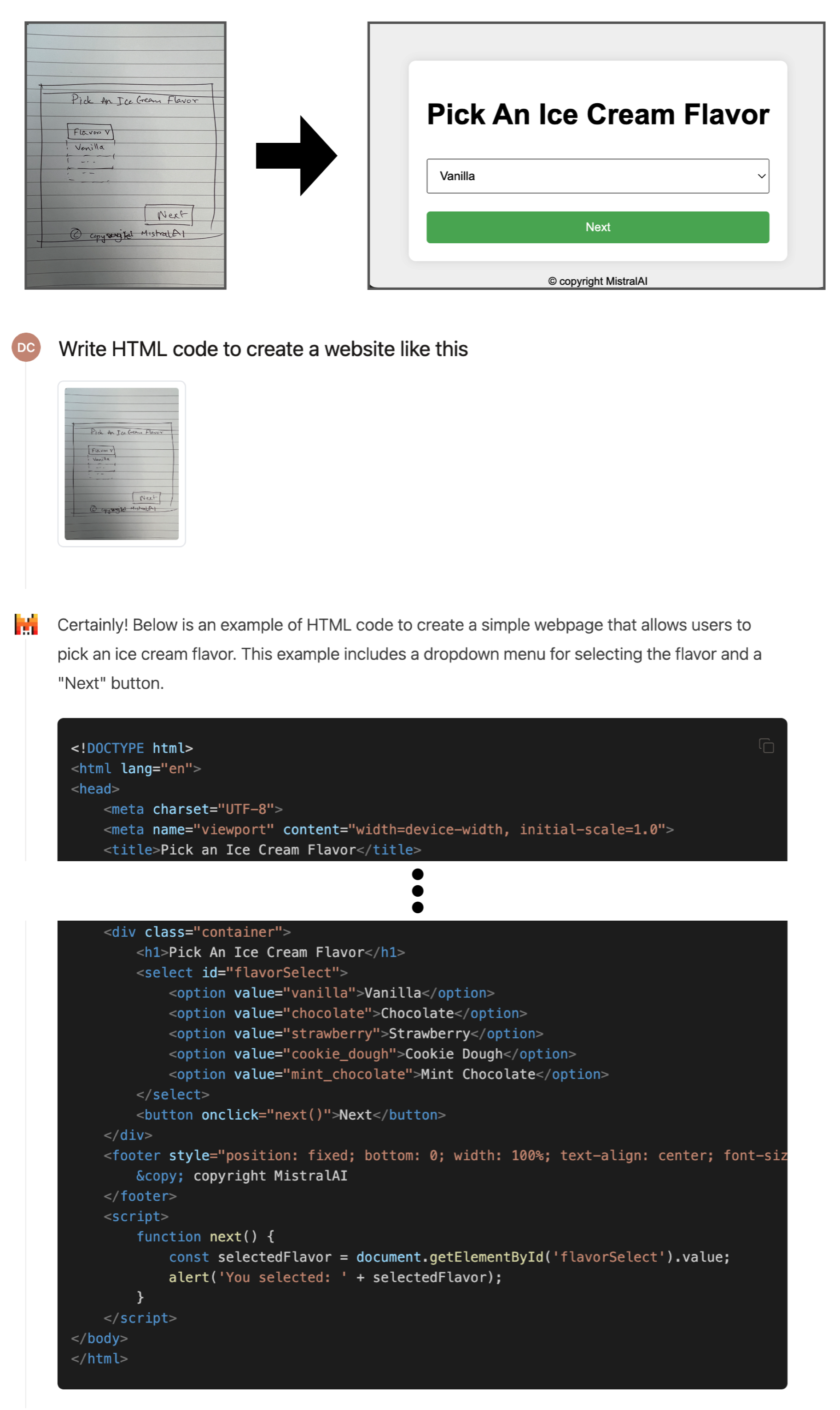}
\caption{\small \textbf{Image to Code.} This demonstration illustrates Pixtral's capability to convert hand-drawn website interfaces into executable HTML code, bringing hand-drawn designs to life as fully functional websites.}
\label{fig:eg_image_to_code}
\end{figure*}

We discuss real world application of Pixtral by looking at some qualitative examples. Specifically, Pixtral can be used for reasoning over complex figures (eg. Fig.~\ref{fig:eg_gdp}), multi-image instruction following (eg. Fig.~\ref{fig:eg_multi_image}), chart understanding and analysis (eg. Fig.~\ref{fig:eg_chart}) and converting image to code (eg. Fig.~\ref{fig:eg_image_to_code}).

\begin{figure}[htbp]
    \centering
    \includegraphics[width=\textwidth]{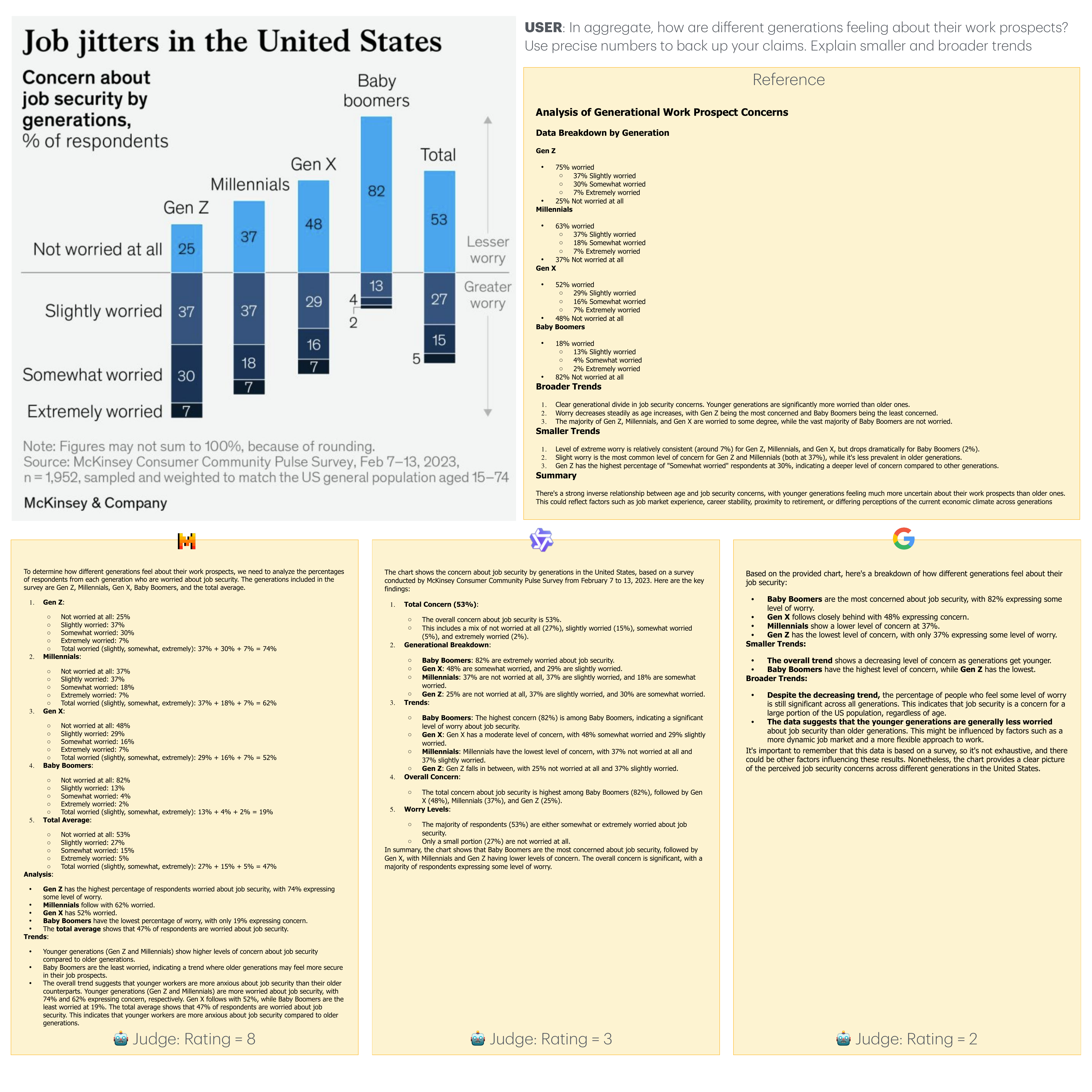}
    \caption{Examples of model responses from Pixral-12B, QwenVL-7B and Gemini-1.5 Flash-8B (0827) LLM-as-a-judge scores. Pixtral's response is complete and accurate, hence getting a rating of 8, while Gemini-Flash-8B extracts wrong information, and QwenVL does not elaborate on trends.}
    \label{fig:model-responses}
\end{figure}

In Fig.~\ref{fig:model-responses}, we compare Pixtral 12B to QwenVL-7B and Gemini-1.5 Flash-8B (0827) on an example from MM-MT-Bench. The example consists of a complex chart on job jitters in the US with an instruction requiring accurate understanding, reasoning and analysis of the chart. Pixtral's response is complete and accurate, hence getting a rating of 8, while Gemini-Flash-8B extracts wrong information, and QwenVL does not elaborate on trends.

\section{Conclusion}
\vspace{-4pt}
\looseness=-1 This paper introduced Pixtral 12B, a state-of-the-art multimodal model that excels in both text-only and multimodal tasks. With a novel architecture featuring a 400M-parameter vision encoder and a 12B-parameter multimodal decoder, Pixtral 12B demonstrates strong performance across various benchmarks, outperforming other open models and matching larger models. Its superior instruction following abilities, support for variable image sizes, and long context window make it highly versatile for complex multimodal applications. Pixtral 12B is released under the Apache 2.0 license.

\section{Contributors}
\vspace{-4pt}
\textbf{Mistral AI Science team }(listed in alphabetical order by last name):

Pravesh Agrawal, Szymon Antoniak, Emma Bou Hanna, Baptiste Bout, Devendra Chaplot, Jessica Chudnovsky, Diogo Costa, Baudouin De Monicault, Saurabh Garg, Theophile Gervet, Soham Ghosh, Amélie Héliou, Paul Jacob, Albert Q. Jiang, Kartik Khandelwal, Timothée Lacroix, Guillaume Lample, Diego Las Casas, Thibaut Lavril, Teven Le Scao, Andy Lo, William Marshall, Louis Martin, Arthur Mensch, Pavankumar Muddireddy, Valera Nemychnikova, Marie Pellat, Patrick Von Platen, Nikhil Raghuraman, Baptiste Rozière, Alexandre Sablayrolles, Lucile Saulnier, Romain Sauvestre, Wendy Shang, Roman Soletskyi, Lawrence Stewart, Pierre Stock, Joachim Studnia, Sandeep Subramanian, Sagar Vaze, Thomas Wang, Sophia Yang.

\subsection*{Acknowledgements}

We extend our thanks to the LMSys team for their assistance in deploying our model in the LLM arena, and the vLLM team for their help in integrating Pixtral 12B into their inference library.

\clearpage
\bibliographystyle{apalike}
\bibliography{ref}

\clearpage
\appendix
\addcontentsline{toc}{section}{Appendix} 
\part{Appendix} 
\parttoc 
\clearpage

\section{Prompts}
\label{sec:appendixA}

Here we open-source the prompts used for evaluations in the main paper.
As discussed in Section~\ref{sec:prompt_selection}, prompts are selected to reproduce reported performance of GPT-4o~\cite{openai2023gpt} and Claude-3.5 Sonnet~\cite{claude3}.

\subsection{MMMU and Mathvista}
\begin{quote}
\small
\begin{verbatim}
Analyze the image and question carefully, using step-by-step reasoning.
First, describe any image provided in detail. Then, present your reasoning. 
And finally your final answer in this format:
Final Answer: <answer>
where <answer> is:
- The single correct letter choice A, B, C, D, E, F, etc. when options are provided.
Only include the letter.
- Your direct answer if no options are given, as a single phrase or number.
- If your answer is a number, only include the number without any unit.
- If your answer is a word or phrase, do not paraphrase or reformat the text 
you see in the image.
- You cannot answer that the question is unanswerable. You must either pick an 
option or provide a direct answer.
IMPORTANT: Remember, to end your answer with Final Answer: <answer>.
\end{verbatim}
\normalsize
\end{quote}

\subsection{ChartQA}
\begin{quote}
\small
\begin{verbatim}
Analyze the image and question carefully, using step-by-step reasoning.
First, describe any image provided in detail. Then, present your reasoning.
And finally your final answer in this format:
Final Answer: <answer>
where <answer> follows the following instructions:
- <answer> should be a single phrase or number.
- <answer> should not paraphrase or reformat the text in the image.
- If <answer> is a ratio, it should be a decimal value like 0.25 instead of 1:4.
- If the question is a Yes/No question, <answer> should be Yes/No.
- If <answer> is a number, it should not contain any units.
- If <answer> is a percentage, it should include a % sign.
- If <answer> is an entity, it should include the full label from the graph.
IMPORTANT: Remember, to end your answer with Final Answer: <answer>.
\end{verbatim}
\normalsize
\end{quote}

\subsection{VQAv2}
\begin{quote}
\small
\begin{verbatim}
- Answer the question using a single word, number, or short phrase. 
  Use as few words as possible.
- If the answer is a number, report it as a number, i.e. 2, not Two, 
  and only include the number without any unit.
- If the question is Yes/No, answer with Yes/No, and nothing else 
  (no likely, unknown, etc.).
- You cannot answer that the question is unanswerable. You must answer.
\end{verbatim}
\normalsize
\end{quote}

\subsection{DocVQA}
\begin{quote}
\small
\begin{verbatim}
Answer the question using a single word or phrase.
\end{verbatim}
\normalsize
\end{quote}

\subsection{MM-MT-Bench Judge Prompt}\label{sec:mm_mt_bench_judge_prompt}
\begin{quote}
\small
\begin{verbatim}
SYSTEM: Please act as an impartial judge and evaluate the quality of the response
provided by an AI assistant to the most recent question given the previous 
conversation as context. Your evaluation should consider correctness and 
helpfulness. You will be given a reference answer and the assistant\'s answer. 
Begin your evaluation by comparing the assistant\'s answer with the reference 
answer. Identify and correct any mistakes. Be as objective as possible. After 
providing your explanation, you must rate the response on a scale of 1 to 10 by 
strictly following this format: "[[rating]]", for example: "Rating: [[5]]".

<|The Start of Conversation with User|>

### User:
<image> Analyze this image.

### Reference answer:
The image consists of ...

### Assistant:
This is an image of...

<|The End of Conversation with User|>\n\n\n
\end{verbatim}
\normalsize
\end{quote}

The history of the conversation is passed to the judge with reference answers as assistant answer (teacher-forcing).

\section{Relative Position Encoding Property of \textsc{RoPE-2D}}
\label{sec:appendixrope}

In this section, we show the relative position encoding property of \textsc{RoPE}-2D. The goal is prove that: $$\inner{\rope(x^{(p,q)}, \Theta)}{\rope(y^{(r,s)}, \Theta)} = \inner{\rope(x^{(p-r,q-s)}, \Theta)}{\rope(y^{(0,0)}, \Theta)}$$ for any feature $x, y \in \mathbb{R}^d$ for all positions $p,r \in \{0 \dots H\}$ and $q,s\in \{0 \dots W\}$. To keep the discussion simple, we will illustrate this property for $d=4$ (the extension to higher dimension is straightforward).

\begin{align*}
    \textsc{RoPE-2D}\left(x^{(p,q)}, \Theta \right) &= \begin{pmatrix}
\cos p \theta_1 & -\sin p\theta_1 & 0 & 0 \\
\sin p \theta_1 & \cos p\theta_1 & 0 & 0 \\
0 & 0 & \cos q\theta_2 & -\sin q\theta_2 \\
0 & 0 & \sin q\theta_2 & \cos q\theta_2\\
\end{pmatrix} \cdot  \begin{pmatrix} x_1 \\ x_2 \\ x_3 \\ x_4
\end{pmatrix} 
\end{align*}

\begin{align*}
    \textsc{RoPE-2D}\left(y^{(r,s)}, \Theta \right) &= \begin{pmatrix}
\cos r \theta_1 & -\sin r\theta_1 & 0 & 0 \\
\sin r \theta_1 & \cos r\theta_1 & 0 & 0 \\
0 & 0 & \cos s\theta_2 & -\sin s\theta_2 \\
0 & 0 & \sin s\theta_2 & \cos s\theta_2\\
\end{pmatrix} \cdot  \begin{pmatrix} y_1 \\ y_2 \\ y_3 \\ y_4
\end{pmatrix} 
\end{align*}

Now, we compute 

\begin{align*}
    & \inner{\rope(x^{(p,q)}, \Theta)}{\rope(y^{(r,s)}, \Theta)} \\
    & \qquad = \begin{pmatrix} x_1 \ \ x_2 
\end{pmatrix} \cdot  \begin{pmatrix}
\cos p \theta_1 & -\sin p\theta_1\\
\sin p \theta_1 & \cos p\theta_1 \\
\end{pmatrix}^T
 \begin{pmatrix}
\cos r \theta_1 & -\sin r\theta_1\\
\sin r \theta_1 & \cos r\theta_1 \\
\end{pmatrix} \cdot 
\begin{pmatrix} y_1 \\ y_2 
\end{pmatrix} \\
& \qquad \qquad  + 
 \begin{pmatrix} x_3 \ \ x_4
\end{pmatrix} \cdot 
 \begin{pmatrix}
\cos q\theta_2 & -\sin q\theta_2 \\
\sin q\theta_2 & \cos q\theta_2
\end{pmatrix}^T
 \begin{pmatrix}
\cos s\theta_2 & -\sin s\theta_2 \\
\sin s\theta_2 & \cos s\theta_2
\end{pmatrix} \cdot  \begin{pmatrix} y_3 \\ y_4
\end{pmatrix} \\
    & \qquad = \begin{pmatrix} x_1 \ \ x_2 
\end{pmatrix} \cdot  \begin{pmatrix}
\cos p \theta_1 \cos r\theta_1 + \sin p \theta_1 \sin r\theta_1 &  - \sin r \theta_1 \cos p\theta_1 + \sin p \theta_1 \cos r\theta_1 \\
\sin r \theta_1 \cos p\theta_1 - \sin p \theta_1 \cos r\theta_1 & \cos p \theta_1 \cos r\theta_1 + \sin p \theta_1 \sin r\theta_1\\
\end{pmatrix}^T
 \cdot 
\begin{pmatrix} y_1 \\ y_2 
\end{pmatrix} \\
& \qquad \qquad  + 
 \begin{pmatrix} x_3 \ \ x_4
\end{pmatrix} \cdot 
 \begin{pmatrix}
\cos q \theta_2 \cos s\theta_2 + \sin q \theta_2 \sin s\theta_2 &  - \sin q \theta_2 \cos s\theta_2 + \sin q \theta_2 \cos s\theta_2\\
\sin q \theta_2 \cos s\theta_2 - \sin q \theta_2 \cos s\theta_2 & \cos q \theta_2 \cos s\theta_2 + \sin q \theta_2 \sin s\theta_2\\
\end{pmatrix} \cdot  \begin{pmatrix} y_3 \\ y_4
\end{pmatrix}\\
    & \qquad = \begin{pmatrix} x_1 \ \ x_2 
\end{pmatrix} \cdot  \begin{pmatrix}
\cos \left( (p -r) \cdot \theta_1 \right) &  - \sin \left( (p-r) \cdot \theta_1 \right) \\
\sin \left( (p-r) \cdot \theta_1 \right) & \cos \left( (p-r) \cdot \theta_1 \right) \\
\end{pmatrix}^T
 \cdot 
\begin{pmatrix} y_1 \\ y_2 
\end{pmatrix} \\
& \qquad \qquad  + 
 \begin{pmatrix} x_3 \ \ x_4
\end{pmatrix} \cdot 
\begin{pmatrix}
\cos \left( (q-s) \cdot \theta_2 \right) &  - \sin \left( (q-s) \cdot \theta_2 \right) \\
\sin \left( (q-s) \cdot \theta_2 \right) & \cos \left( (q-s) \cdot \theta_2 \right) \\
\end{pmatrix}^T \cdot  \begin{pmatrix} y_3 \\ y_4
\end{pmatrix} \\
& \qquad = \inner{\rope(y^{(p-r,q-s)}, \Theta)}{\rope(y^{(0,0)}, \Theta)}
\end{align*}

\section{Flexible Parsing Settings}
\label{sec_app:parsing_ablations}

In Section~\ref{sec:parsing_ablation}, we introduce three `parsing levels' which evaluate models under progressively looser constraints.
While common evaluation metrics reward only exactly the answer format in the ground truth annotation, we seek to relax these requirements and investigate how model performance varies.

\textbf{Baseline:} This setting requires exact following of prompt instructions, with model responses ending in the string \texttt{"Final Answer: <ANSWER>"}.

\textbf{Flexible Parsing Level 1:} This setting also catches cases where the model ends responses with \texttt{"Answer: <ANSWER>"}.

\textbf{Flexible Parsing Level 2:} Here we additionally catch cases where the model has added extra markdown formatting. We strip markdown such as: \texttt{"**Answer**", "**Answer:**", "*Answer: <ANSWER>*"}. 
We find such formatting to be particularly prevalent in Llama-3.2 models~\cite{dubey2024llama3}.

\textbf{Flexible Parsing Level 3:} This is the most generous evaluation setting. 
Here we mark a response as correct if the ground truth answer appears \textit{anywhere} in the model's response. 
For single letter answers, we search the response for \texttt{"is <A>", "are <A>", "<A>"}.
For single number responses, we search the response for the number both with and without commas. 

We highlight that Flexible Parsing Level 3 is intended to serve as an upper bound, as it may mark incorrect answers as correct. 

\section{Robustness to prompting}
\label{sec_app:average_results}

\subsection{Llama-Specific Prompts}

In Section~\ref{sec:main_results}, we evaluate all models with a common prompt, which allowed us to reproduce the reported figures of GPT-4o~\cite{openai2023gpt} and Claude-3.5 Sonnet~\cite{claude3}. This prompt requires models to end responses with \texttt{"Final Answer: <ANSWER>"} (see Appendix~\ref{sec:appendixA} for full prompts).

However, when evaluating Llama-3.2 models~\cite{dubey2024llama3}, we found that this model family defaults to responding with \texttt{"**Answer:** <ANSWER>"} (\textit{i.e.}, with markdown formatting and omission of \textit{`Final'}, despite the explicit instruction). 
While the performance degradation due to regex mismatches is mitigated through our flexible parsing strategy (see Section~\ref{sec:parsing_ablation}), we found that Llama-3.2 models performed substantially better when the \textit{prompt} specifically asks for \texttt{"**Answer:** <ANSWER>"} (\textit{i.e.}, respecting its default output format).

In Table~\ref{tab:llama_prompts}, we show the results for models both with the default prompts from Appendix~\ref{sec:appendixA}, and with the Llama-specific prompts (all evaluated under the \textit{Exact Match} metric).
We show that the Llama-specific prompt substantially improves the performance of Llama-3.2 models, particularly for the 11B variant, with over 15\% jumps on both Mathvista and MMMU. 
We further note that Pixtral performance is stable across prompts, and leads the 11B variant by a substantial margin. 

\definecolor{flexOne}{RGB}{199,232,229}   
\definecolor{flexTwo}{RGB}{239,245,214}   
\definecolor{flexThree}{RGB}{253,224,175} 
\begin{table}[htbp]
\centering
\small
\begin{tabular}{@{}llcccc@{}}
\toprule
\rowcolor{white}
& & \multicolumn{1}{c}{\cellcolor{flexOne}\textbf{Mathvista}} & \multicolumn{1}{c}{\cellcolor{flexTwo}\textbf{MMMU}} & \multicolumn{1}{c}{\cellcolor{flexThree}\textbf{ChartQA}} \\
\cmidrule(lr){3-3} \cmidrule(lr){4-4} \cmidrule(lr){5-5}
& & Exact Match & Exact Match & Exact Match \\
\midrule
\multirow{2}{*}{Llama-3.2 11B~\cite{dubey2024llama3}} & Default prompt & 24.3 & 23.0 & 14.8 \\
& Llama-specific prompt & 41.6 & 41.9 & 33.7 \\
\midrule
\multirow{2}{*}{Llama-3.2 90B~\cite{dubey2024llama3}} & Default prompt & 49.1 & 53.7 & 33.8 \\
& Llama-specific prompt & 57.6 & \textbf{58.6} & 34.8 \\
\midrule
\multirow{2}{*}{Qwen2-VL 7B~\cite{wang2024qwen2vlenhancingvisionlanguagemodels}} & Default prompt & 53.7 & 48.1 & 41.2 \\
& Llama-specific prompt & 52.6 & 47.4 & 74.0 \\
\midrule
\multirow{2}{*}{Pixtral 12B} & Default prompt & \textbf{58.3} & 52.0 & 81.8 \\
& Llama-specific prompt & 57.7 & 50.8 & \textbf{83.8} \\
\bottomrule
\end{tabular}
\vspace{1mm}
\caption{\textbf{Evaluation with Llama-specific prompts.}
We re-evaluate models with a prompt tailored towards the Llama-3.2 model family~\cite{dubey2024llama3}.
We find that this substantially improves the performance of the 11B variant of the model.
Pixtral 12B reports stable performance across both prompts, and maintains a substantial lead over Llama-3.2 11B and Qwen2-VL 7B.
}
\label{tab:llama_prompts}
\end{table}

\subsection{Average performance across prompts}

Here we report average results across a number of prompts.
We task Mistral Large v2 with creating 10 versions of the prompt used in the main paper (see Appendix~\ref{sec:appendixA}), with varied wording while keeping instructions explicit. 
As prior works suffer under stricter parsing constraints, all models are evaluated under `Flexible Parsing Level 3' for this experiment (see Section~\ref{sec:parsing_ablation} and Appendix~\ref{sec_app:parsing_ablations}).

We find that the trends follow those from the main paper, with Pixtral outperforming models of comparable size, and surpassing Llama-3.2 90B~\cite{dubey2024llama3} on Mathvista and ChartQA.
Pixtral also typically displays lower variance in performance between prompts (shown in gray).

\definecolor{flexOne}{RGB}{199,232,229}   
\definecolor{flexTwo}{RGB}{239,245,214}   
\definecolor{flexThree}{RGB}{253,224,175} 
\begin{table}[htbp]
\centering
\small
\begin{tabular}{@{}lccc@{}}
\toprule
\rowcolor{white}
& \multicolumn{1}{c}{\cellcolor{flexOne}\textbf{Mathvista}} & \multicolumn{1}{c}{\cellcolor{flexTwo}\textbf{MMMU}} & \multicolumn{1}{c}{\cellcolor{flexThree}\textbf{ChartQA}} \\
\cmidrule(lr){2-2} \cmidrule(lr){3-3} \cmidrule(lr){4-4}
 & Flexible Level 3 & Flexible Level 3 & Flexible Level 3 \\
\midrule
Llama-3.2 11B~\cite{dubey2024llama3} & 42.1 {\scriptsize\textcolor{gray}{($\pm$1.9)}} & 45.3 {\scriptsize\textcolor{gray}{($\pm$1.0)}} & 77.2 {\scriptsize\textcolor{gray}{($\pm$0.8)}} \\
Llama-3.2 90B~\cite{dubey2024llama3} & 56.0 {\scriptsize\textcolor{gray}{($\pm$1.5)}} & \textbf{56.7} {\scriptsize\textcolor{gray}{($\pm$0.5)}} & 80.1 {\scriptsize\textcolor{gray}{($\pm$0.5)}} \\
Qwen2-VL 7B~\cite{wang2024qwen2vlenhancingvisionlanguagemodels} & 53.7 {\scriptsize\textcolor{gray}{($\pm$2.1)}} & 46.9 {\scriptsize\textcolor{gray}{($\pm$1.9)}} & 77.0 {\scriptsize\textcolor{gray}{($\pm$0.8)}} \\
Pixtral 12B & \textbf{56.4} {\scriptsize\textcolor{gray}{($\pm$1.0)}} & 49.5 {\scriptsize\textcolor{gray}{($\pm$1.5)}} & \textbf{83.8} {\scriptsize\textcolor{gray}{($\pm$0.4)}} \\
\bottomrule
\end{tabular}
\vspace{1mm}
\caption{\textbf{Average multimodal performance across prompts}.
We evaluate models with 10 different prompts, reporting the mean performance, and standard deviations in gray.
The trends follow those in the main paper, with Pixtral outperforming open-source models of a comparable size.
All models are evaluated with `Flexible Level 3' parsing (see Section~\ref{sec:parsing_ablation})
}
\label{tab:average_prompts}
\end{table}

\section{Reproducing Reported Numbers}
\label{sec_app:reproducing_reported}
In Section~\ref{sec:main_results} we re-evaluate all models under a common and rigorous protocol. 
All models are evaluated under the same evaluation metric and with the same prompt, in such a way that frontier models achieve their reported performance. 

Under this common protocol, we found some models substantially underperformed their reported figures.
Here, we document the steps required to recover the reported figures of open models, by tuning the evaluation prompt and metric to each model in turn.
All results are shown in Table~\ref{tab:reproducing_reported}.

\subsection{Summary}

Our analysis indicates that frontier models, and even smaller closed-source models, are able to recover or exceed their reported figures under the common protocol discussed in Section~\ref{sec:main_results}.
This is achieved through precise following of instructions in the `Explicit' prompts (see Appendix~\ref{sec:appendixA}).

Smaller, open-source models typically require some degree of prompt tuning and/or adjustment of the evaluation metric, targeted towards the model, to recover reported performance.
With such interventions, we are generally able to recover or exceed reported figures.

Pixtal 12B, like closed and leading models, is able to follow prompt instructions to report strong performance without targeted interventions.
This is substantiated in robust performance across  prompts (see Appendix~\ref{sec_app:average_results}), as well as strong performance in both LMSys Vision Arena and MM-MT-Bench (see Section~\ref{sec:main_results}).

\subsection{Closed models: Claude-3 Haiku and Gemini-Flash-8B}
\label{sec_app:repro_closed}

We find we the standardized evaluation protocol roughly matches or exceeds reported figures, with a small gain achieved through flexible parsing.
The only exception is for Claude Haiku~\cite{claude3} on ChartQA, where Flexible Parsing Level 3 is required to approach reported performance.

\subsection{Qwen2-VL 7B}
\label{sec_app:repro_qwen2}

We first simplify the prompt into a one-line instruction, similar to the training set of ChartQA.
Next, we provide different prompts depending on the answer format expected. 
For instance, if the answer is a floating point number, we specify \texttt{"Answer with a two decimal place floating point"}, with analogous prompts for integer and multiple-choice questions. 
We found that providing a single, unified prompt with all format specifications (as in the prompts in Appendix~\ref{sec:appendixA}) reduces performance.

\subsection{Llama-3.2}
\label{sec_app:repro_llama}
We find that these models default to responses with markdown formatting such as: \texttt{"**Answer**", "**Answer:**", "*Answer: <ANSWER>*"}.
We find substantial improvement by changing the `Explicit' prompt to request this format (see Appendix~\ref{sec_app:average_results}).
These models then recover their reported performance after evaluating with Flexible Level 3. 

When evaluating Llama-3.2 90B on DocVQA , many generations are of the form `\texttt{The answer is <ANSWER>}', which is penalized by the ANLS metric. 
We strip such prefixes, and this improves DocVQA by $+4.8$.

\subsection{Llava-OneVision 72B}
\label{sec_app:repro_llava}

Similarly to Qwen2-7B~\cite{wang2024qwen2vlenhancingvisionlanguagemodels}, we first simplify the prompt into a one-line instruction and provide different prompts depending on the answer format expected. 
We found that providing a single, unified prompt with all format specifications reduces performance.

\subsection{Molmo}
\label{sec_app:repro_molmo}

Similarly to Qwen2-7B~\cite{wang2024qwen2vlenhancingvisionlanguagemodels} and Llava-Onevision 7B~\cite{li2024llava}, we first simplify the prompt into a one-line instruction, and provide different prompts depending on the answer format expected. 
Furthermore, similarly to the intervention for Llama-3.2~\cite{dubey2024llama3}, we reformat the prompt and relax the evaluation metrics.
Molmo models default to ending long responses with \texttt{\textbackslash n\textbackslash n<ANSWER>}.
In long-answer cases, we adjust the evaluation metric to capture this.

For VQAv2, we apply custom post-processing filters, such as remapping textual output of numerical answers to the integer digits (\textit{e.g.} \texttt{Two} to \texttt{2}).

\begin{table}
\small
\renewcommand{\arraystretch}{1.2}
\resizebox{\linewidth}{!}{%
\begin{tabular}{@{}l*{7}{c}@{}}
\toprule
& \textbf{Mathvista} & \textbf{MMMU} & \textbf{ChartQA} & \textbf{DocVQA} & \textbf{VQAv2} & \textbf{MM-MT-Bench} & \textbf{LMSys-Vision} \\[-2pt]
& \textcolor{gray}{\scriptsize CoT} & \textcolor{gray}{\scriptsize CoT} & \textcolor{gray}{\scriptsize CoT} & \textcolor{gray}{\scriptsize ANLS} & \textcolor{gray}{\scriptsize VQA Match} & \textcolor{gray}{\scriptsize GPT-4o Judge} & \textcolor{gray}{\scriptsize (Oct '24)}  \\
\midrule
\rowcolor{flexThree}
\textbf{Pixtral 12B} & 58.3 & 52.0 & 81.8 & 90.7 & 78.6 & 6.05 & 1076 \\
\midrule
\textbf{Qwen-2-VL 7B}~\cite{wang2024qwen2vlenhancingvisionlanguagemodels} & & & & & & & \\
\midrule
Measured (Exact Match) & 53.7 & 48.1 & 41.2 & 94.5 & 75.9 & 5.45 & \multirow{3}{*}{1040} \\
Measured (Custom evaluation, see Section~\ref{sec_app:repro_qwen2}) & 63.7 & 50.6 & 83.4 & 94.5 & 82.1 & - & \\
\textcolor{gray}{Reported} & \textcolor{gray}{58.2} & \textcolor{gray}{54.1} & \textcolor{gray}{83.0} & \textcolor{gray}{94.5} & \textcolor{gray}{-} & \textcolor{gray}{-} & \\
\midrule
\textbf{Llama-3.2 11B}~\cite{dubey2024llama3} & & & & & & & \\
\midrule
Measured (Exact Match)  & 24.3 & 23.0 & 14.8 & 91.1 & 67.1 & 4.79 & \multirow{3}{*}{1032} \\
Measured (Custom evaluation, see Section~\ref{sec_app:repro_llama})  & 47.9 & 46.6 & 78.5 & 91.1 & 67.1 & - & \\
\textcolor{gray}{Reported} & \textcolor{gray}{51.5} & \textcolor{gray}{50.7} & \textcolor{gray}{83.4} & \textcolor{gray}{88.4} & \textcolor{gray}{75.2} & \textcolor{gray}{-} & \\
\midrule
\textbf{Molmo-D 7B}~\cite{deitke2024molmo}  & & & & & & & \\
\midrule
Measured (Exact Match)  & 12.3 & 24.3 & 27.0 & 72.2 & 57.1 & 3.72 & \multirow{3}{*}{--} \\
Measured (Custom evaluation, see Section~\ref{sec_app:repro_molmo})  & 43.2 & 47.0 & 76.7 & 72.2 & 70.0 & - & \\
\textcolor{gray}{Reported} & \textcolor{gray}{51.6} & \textcolor{gray}{45.3} & \textcolor{gray}{84.1} & \textcolor{gray}{92.2} & \textcolor{gray}{85.6} & \textcolor{gray}{-} & \\
\midrule
\textbf{LLaVA-OneVision 7B}~\cite{li2024llava} & & & & & & & \\
\midrule
Measured (Exact Match) & 36.1 & 45.1 & 67.2 & 90.5 & 78.4 & 4.12 & \multirow{3}{*}{--} \\
Measured (Custom evaluation, see Section~\ref{sec_app:repro_llava}) & 63.1 & 48.1 & 80.2 & 90.5 & 83.7 & - & \\
\textcolor{gray}{Reported} & \textcolor{gray}{63.2} & \textcolor{gray}{48.8} & \textcolor{gray}{80.0} & \textcolor{gray}{87.5} & \textcolor{gray}{-} & \textcolor{gray}{-} & \\
\midrule
\textbf{Molmo 72B}~\cite{deitke2024molmo}  & & & & & & & -- \\
\midrule
Measured (Exact Match) & 52.2 & 52.7 & 75.6 & 86.5 & 75.2 & 3.51 & \multirow{3}{*}{--} \\
Measured (Custom evaluation, see Section~\ref{sec_app:repro_molmo}) & 61.3 & 52.9 & 82.3 & 86.5 & 75.5 & - & \\
\textcolor{gray}{Reported} & \textcolor{gray}{58.6} & \textcolor{gray}{54.1} & \textcolor{gray}{87.3} & \textcolor{gray}{93.5} & \textcolor{gray}{86.5} & \textcolor{gray}{-} & \\
\midrule
\textbf{Llama-3.2 90B}~\cite{dubey2024llama3} & & & & & & & \\
\midrule
Measured (Exact Match) & 49.1 & 53.7 & 33.8 & 85.7 & 67.0 & 5.50 & \multirow{3}{*}{1071} \\
Measured (Custom evaluation, see Section~\ref{sec_app:repro_llama}) & 57.5 & 60.2 & 91.7 & 91.5 & 67.0 & - & \\
\textcolor{gray}{Reported} & \textcolor{gray}{57.3} & \textcolor{gray}{60.3} & \textcolor{gray}{85.5} & \textcolor{gray}{90.1} & \textcolor{gray}{78.1} & \textcolor{gray}{-} & \\
\midrule
\textbf{Claude-3 Haiku}~\cite{claude3} & & & & & & & \\
\midrule
Measured (Exact Match) & 44.8 & 50.4 & 69.6 & 74.6 & 68.4 & 5.46 & \multirow{3}{*}{1000} \\
Measured (Custom evaluation, see Section~\ref{sec_app:repro_closed}) & 44.8 & 51.3 & 79.8 & 74.6 & 68.4 & - & \\
\textcolor{gray}{Reported} & \textcolor{gray}{46.4} & \textcolor{gray}{50.2} & \textcolor{gray}{81.7} & \textcolor{gray}{88.8} & \textcolor{gray}{-} & \textcolor{gray}{-} & \\
\midrule
\textbf{Gemini-1.5-Flash 8B}\textcolor{gray}{\scriptsize (0827)}~\cite{reid2024gemini} & & & & & & & \\
\midrule
Measured (Exact Match) & 56.9 & 50.7 & 78.0 & 79.5 & 65.5 & 5.93 & \multirow{3}{*}{1111} \\
Measured (Custom evaluation, see Section~\ref{sec_app:repro_closed}) & 57.1 & 50.7 & 78.2 & 79.5 & 69.2 & - & \\
\textcolor{gray}{Reported} & \textcolor{gray}{-} & \textcolor{gray}{50.3} & \textcolor{gray}{-} & \textcolor{gray}{73.6} & \textcolor{gray}{-} & \textcolor{gray}{-} & \\
\bottomrule
\end{tabular}%
}
\vspace{1mm}
\caption{\textbf{Reproducing the reported performance of prior models.}
In Table~\ref{tab:oss_comparisons} we conduct fair re-evaluation of all models through the same evaluation harness, with the same prompt and metric.
Here, we endeavour to recover the reported performance of all models by tuning evaluation settings towards individual models. 
We highlight that Pixtral 12B, like strong closed-source models (\textit{e.g.} Gemini-1.5-Flash 8B~\cite{reid2024gemini} and Claude-3 Haiku~\cite{claude3}) is able reports strong performance without such interventions. 
}
\label{tab:reproducing_reported}
\end{table}

\end{document}